\documentclass{article}

\usepackage{microtype}
\usepackage{graphicx}
\usepackage{subcaption}

\usepackage{booktabs} %

\usepackage{hyperref}

\usepackage[accepted]{icml2024}

\usepackage{amsmath}
\usepackage{amssymb}
\usepackage{mathtools}
\usepackage{amsthm}

\usepackage[capitalize,noabbrev]{cleveref}

\theoremstyle{plain}
\newtheorem{theorem}{Theorem}[section]
\newtheorem{proposition}[theorem]{Proposition}
\newtheorem{lemma}[theorem]{Lemma}
\newtheorem{corollary}[theorem]{Corollary}
\theoremstyle{definition}
\newtheorem{definition}[theorem]{Definition}
\newtheorem{assumption}[theorem]{Assumption}
\theoremstyle{remark}
\newtheorem{remark}[theorem]{Remark}
\usepackage{mathtools}
\usepackage{amssymb}
\usepackage{latexsym}
\usepackage{dsfont}
\usepackage{mathrsfs}
\usepackage{amssymb}
\usepackage{amsfonts}
\usepackage{bm}
\usepackage{xspace}
\usepackage{amsthm}

\ifx\BlackBox\undefined
\newcommand{\BlackBox}{\rule{1.5ex}{1.5ex}}  
\fi

\ifx\QED\undefined
\def\QED{~\rule[-1pt]{5pt}{5pt}\par\medskip}
\fi

\ifx\proof\undefined
\newenvironment{proof}{\par\noindent{\em Proof:\ }}{\hfill\BlackBox\\[.0mm]}
\fi

\ifx\theorem\undefined
\newtheorem{theorem}{Theorem}[section]
\fi

\ifx\example\undefined
\newtheorem{example}{Example}[section]
\fi

\ifx\property\undefined
\newtheorem{property}{Property}
\fi

\ifx\lemma\undefined
\newtheorem{lemma}{Lemma}[section]
\fi

\ifx\proposition\undefined
\newtheorem{proposition}{Proposition}[section]
\fi

\ifx\remark\undefined
\newtheorem{remark}{Remark}
\fi

\ifx\corollary\undefined
\newtheorem{corollary}{Corollary}[section]
\fi

\ifx\definition\undefined
\newtheorem{definition}{Definition}[section]
\fi

\ifx\conjecture\undefined
\newtheorem{conjecture}{Conjecture}[section]
\fi

\ifx\axiom\undefined
\newtheorem{axiom}[theorem]{Axiom}
\fi

\ifx\claim\undefined
\newtheorem{claim}[theorem]{Claim}
\fi

\ifx\assumption\undefined
\newtheorem{assumption}{Assumption}
\fi

\ifx\problem\undefined
\newtheorem{problem}{Problem}[section]
\fi

\ifx\fact\undefined
\newtheorem{fact}{Fact}
\fi

\newcommand{\rbr}[1]{\left(#1\right)}
\newcommand{\sbr}[1]{\left[#1\right]}

\newcommand{\ours}
{\textsc{Medusa}\xspace}

\usepackage[textsize=tiny]{todonotes}

\icmltitlerunning{\ours: Simple LLM Inference Acceleration Framework with Multiple Decoding Heads}

\begin{document}

\twocolumn[
\icmltitle{\ours: Simple LLM Inference Acceleration Framework with Multiple Decoding Heads}

\icmlsetsymbol{equal}{*}

\begin{icmlauthorlist}
\icmlauthor{Tianle Cai}{equal,princeton,together}
\icmlauthor{Yuhong Li}{equal,uiuc}
\icmlauthor{Zhengyang Geng}{cmu}
\icmlauthor{Hongwu Peng}{uconn}
\icmlauthor{Jason D. Lee}{princeton}
\icmlauthor{Deming Chen}{uiuc}
\icmlauthor{Tri Dao}{princeton,together}
\end{icmlauthorlist}

\icmlaffiliation{princeton}{Princeton University}
\icmlaffiliation{together}{Together AI}
\icmlaffiliation{uiuc}{University of Illinois Urbana-Champaign}
\icmlaffiliation{cmu}{Carnegie Mellon University}
\icmlaffiliation{uconn}{University of Connecticut}

\icmlcorrespondingauthor{Tianle Cai}{tianle.cai@princeton.edu}
\icmlcorrespondingauthor{Yuhong Li}{leeyh@illinois.edu}

\icmlkeywords{Machine Learning, ICML}

\vskip 0.3in
]

\printAffiliationsAndNotice{\icmlEqualContribution} %

\begin{abstract}
\textcolor{black}{
Large Language Models (LLMs) employ auto-regressive decoding that requires sequential computation, with each step reliant on the previous one's output. This creates a bottleneck as each step necessitates moving the full model parameters from High-Bandwidth Memory (HBM) to the accelerator's cache.
}
While methods such as speculative decoding have been suggested to address this issue, their implementation is impeded by the challenges associated with acquiring and maintaining a separate draft model. 
In this paper, we present \ours, an efficient method that augments LLM inference by adding extra decoding heads to predict multiple subsequent tokens in parallel. Using a \emph{tree-based attention mechanism}, \ours constructs multiple candidate continuations and verifies them simultaneously in each decoding step. By leveraging parallel processing, 
\ours substantially \textcolor{black}{reduces} the number of decoding steps required.
We present two levels of fine-tuning procedures for \ours to meet the needs of different use cases: \textbf{\ours-1}: \ours is directly fine-tuned on top of a \emph{frozen} backbone LLM, enabling lossless inference acceleration. \textbf{\ours-2}: \ours is fine-tuned together with the backbone LLM, enabling better prediction accuracy of \ours heads and higher speedup but needing a special training recipe that preserves the model's capabilities. 
Moreover, we propose several extensions that improve or expand the utility of \ours, including a \emph{self-distillation} to handle situations where no training data is available and a \emph{typical acceptance scheme} to boost the acceptance rate while maintaining generation quality.
We evaluate \ours on models of various sizes and training procedures. Our experiments demonstrate that \ours-1 can achieve over 2.2$\times$ speedup without compromising generation quality, while \ours-2 further improves the speedup to 2.3-2.8$\times$.
\end{abstract}

\section{Introduction}
The recent advancements in Large Language Models (LLMs) have demonstrated that the quality of language generation significantly improves with an increase in model size, reaching billions of parameters~\citep{brown2020language,chowdhery2022palm,zhang2022opt,hoffmann2022training,openai2023gpt4,google2023palm2,touvron2023llama}. However, this growth has led to an increase in \emph{inference latency}, which poses a significant challenge in practical applications. From a system perspective, LLM inference is predominantly memory-bandwidth-bound~\citep{shazeer2019fast,kim2023squeezellm}, with the main latency bottleneck stemming from accelerators' memory bandwidth rather than arithmetic computations. This bottleneck is inherent to the sequential nature of auto-regressive decoding, where each forward pass requires transferring the complete model parameters from High-Bandwidth Memory (HBM) to the accelerator's cache. This process, which generates only a single token, underutilizes the arithmetic computation potential of modern accelerators, leading to inefficiency.

To address this, one approach to speed up LLM inference involves \emph{increasing the arithmetic intensity} (the ratio of total floating-point operations (FLOPs) to total data movement) of the decoding process and \emph{reducing the number of decoding steps}. In line with this idea, speculative decoding has been proposed~\citep{leviathan2022fast,chen2023accelerating,xia2023speculative,miao2023specinfer}. This method uses a smaller draft model to generate a token sequence, which is then refined by the original, larger model for acceptable continuation. However, obtaining an appropriate draft model remains challenging, and it's even harder to integrate the draft model into a distributed system~\citep{chen2023accelerating}.

Instead of using a separate draft model to sequentially generate candidate outputs, in this paper, we revisit and refine the concept of using multiple decoding heads on top of the backbone model to expedite inference~\citep{stern2018blockwise}. We find that when applied effectively, this technique can overcome the challenges of speculative decoding, allowing for seamless integration into existing LLM systems. Specifically, we introduce \ours, a method that enhances LLM inference by integrating additional decoding heads to concurrently predict multiple tokens. These heads are fine-tuned in a \emph{parameter-efficient} manner and can be added to any existing model. With no requirement for a draft model, \ours offers easy integration into current LLM systems, including those in distributed environments, ensuring a user-friendly experience.

We further enhance \ours with two key insights. Firstly, the current approach of generating a single candidate continuation at each decoding step leads to inefficient use of computational resources. To address this, we propose generating multiple candidate continuations using the \ours heads and verifying them concurrently through a simple adjustment to the attention mask. 
\textcolor{black}{Secondly, we can reuse the rejection sampling scheme as used in speculative decoding~\cite{leviathan2022fast,chen2023accelerating} to generate consistent responses with the same distribution as the original model. However, it cannot further enhance the acceleration rate.}
Alternatively, we introduce a \emph{typical acceptance} scheme that selects \emph{reasonable} candidates from the \ours head outputs. We use temperature as a threshold to manage deviation from the original model’s predictions, providing an efficient alternative to the rejection sampling method. 
\textcolor{black}{Our results suggest that the proposed typical acceptance scheme can accelerate the decoding speed further while maintaining a similar generation quality.}

To equip LLMs with predictive \ours heads, we propose two distinct fine-tuning procedures tailored to various scenarios. For situations with limited computational resources or when the objective is to incorporate \ours into an existing model without affecting its performance, we recommend \ours-1. This method requires minimal memory and can be further optimized with quantization techniques akin to those in QLoRA~\citep{dettmers2023qlora}, without compromising the generation quality due to the fixed backbone model. However, in \ours-1, the full potential of the backbone model is not utilized. We can further fine-tune it to enhance the prediction accuracy of \ours heads, which can directly lead to a greater speedup.
Therefore, we introduce \ours-2, which is suitable for scenarios with ample computational resources or for direct Supervised Fine-Tuning (SFT) from a base model. 
The key to \ours-2 is a training protocol that enables joint training of the \ours heads and the backbone model without compromising the model’s next-token prediction capability and output quality.
We propose different strategies for obtaining the training datasets depending on the model’s training recipe and dataset availability. When the model is fine-tuned on a public dataset, it can be directly used for \ours. If the dataset is unavailable or the model underwent a Reinforcement Learning with Human Feedback (RLHF)~\citep{ouyang2022training} process, we suggest a self-distillation approach to generate a training dataset for the \ours heads.

Our experiments primarily focus on scenarios with a batch size of one, which is representative of the use case where LLMs are locally hosted for personal use. We test \ours on models of varying sizes and training settings, including Vicuna-7B, 13B (trained with a public dataset), Vicuna-33B~\citep{vicuna2023} (trained with a private dataset\footnote{Upon contacting the authors, this version is experimental and used some different data than Vicuna 7B and 13B.}), and Zephyr-7B (trained with both supervised fine-tuning and alignment). \ours can achieve a speedup of 2.3 to 2.8 times across different prompt types without compromising on the quality of generation.

\begin{figure}[h!]
    \centering
    \includegraphics[width=0.45\textwidth]{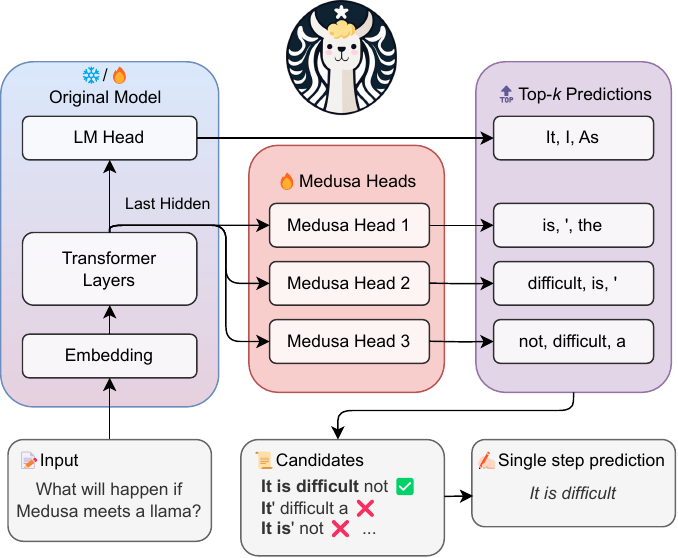}
    \caption{
    \ours introduces \emph{multiple heads} on top of the last hidden states of the LLM, enabling the prediction of several subsequent tokens in parallel (Section~\ref{sec:medusa_heads}). 
    During inference, each head generates multiple top predictions for its designated position. 
    These predictions are assembled into candidates, which are processed in parallel using a \emph{tree-based attention} mechanism (Section~\ref{sec:tree_attention}). The final step is to verify the candidates and accept a continuation. Besides the standard rejection sampling scheme, a \emph{typical acceptance} scheme (Section~\ref{sec:typical_acceptance}) can also be used here to select reasonable continuations, and the \emph{longest accepted candidate prefix} will be used for the next decoding phase. 
    }
    \label{fig:pipeline}
\end{figure}
\section{Methodology}
\ours follows the same framework as speculative decoding, where each decoding step primarily consists of three substeps: (1) generating candidates, (2) processing candidates, and (3) accepting candidates. For \ours, (1) is achieved by \ours heads, (2) is realized by tree attention, and since \ours heads are on top of the original model, the logits calculated in (2) can be used for substep (1) for the next decoding step. The final step (3) can be realized by either rejection sampling~\citep{leviathan2022fast,chen2023accelerating} or typical acceptance (Section~\ref{sec:typical_acceptance}). The overall pipeline is illustrated in Figure~\ref{fig:pipeline}.

In this section, we first introduce the key components of \ours, including \ours heads, and tree attention. Then, we present two levels of fine-tuning procedures for \ours to meet the needs of different use cases. Finally, we propose two extensions to \ours, including self-distillation and typical acceptance, to handle situations where no training data is available for \ours and to improve the efficiency of the decoding process, respectively.
\subsection{Key Components}
\subsubsection{\ours Heads}
\label{sec:medusa_heads}
In speculative decoding, subsequent tokens are predicted by an auxiliary draft model. This draft model must be small yet effective enough to generate continuations that the original model will accept. Fulfilling these requirements is a challenging task, and existing approaches~\citep{spector2023accelerating,miao2023specinfer} often resort to separately \emph{pre-training} a smaller model. This pre-training process demands substantial additional computational resources. For example, in \citep{miao2023specinfer}, a reported 275 NVIDIA A100 GPU hours were used. Additionally, separate pre-training can potentially create a distribution shift between the draft model and the original model, leading to continuations that the original model may not favor. \citet{chen2023accelerating} have also highlighted the complexities of serving multiple models in a distributed environment.

\textcolor{black}{To streamline and democratize the acceleration of LLM inference, we take inspiration from \citet{stern2018blockwise}, which utilizes parallel decoding for tasks such as machine translation and image super-resolution. \ours heads}
 are additional decoding heads appended to the last hidden states of the original model. Specifically, given the original model's last hidden states $h_t$ at position $t$, we add $K$ decoding heads to $h_t$. The $k$-th head is used to predict the token in the $(t+k+1)$-th position of the next tokens (the original language model head is used to predict the $(t+1)$-th position). The prediction of the $k$-th head is denoted as $p_t^{(k)}$, representing a distribution over the vocabulary, while the prediction of the original model is denoted as $p_t^{(0)}$. Following the approach of \citet{stern2018blockwise}, we utilize a single layer of feed-forward network with a residual connection for each head. We find that this simple design is sufficient to achieve satisfactory performance. The definition of the $k$-th head is outlined as:

\begin{align*}
p_t^{(k)} = \text{softmax}\left(W_2^{(k)} \cdot \left(\text{SiLU}(W_1^{(k)} \cdot h_t)+h_t\right)\right),\\
\text{where } W_2^{(k)}\in\mathbb{R}^{d\times V}, W_1^{(k)}\in\mathbb{R}^{d\times d}.
\end{align*}

\textcolor{black}{$d$ is the output dimension of the LLM's last hidden layer and $V$ is the vocabulary size.}
\textcolor{black}{We initialize $W_2^{(k)}$ identically to the original language model head, and $W_1^{(k)}$ to zero.}
This aligns the initial prediction of \ours heads with that of the original model. The SiLU activation function~\citep{elfwing2017sigmoidweighted} is employed following the Llama models~\citep{touvron2023llama}.

Unlike a draft model, \ours heads are trained in conjunction with the original backbone model, which can remain \emph{frozen} during training (\ours-1) or be trained together (\ours-2). This method allows for fine-tuning large models even on a single GPU, taking advantage of the powerful base model's learned representations. Furthermore, it ensures that the distribution of the \ours heads aligns with that of the original model, thereby mitigating the distribution shift problem. Additionally, since the new heads consist of just a single layer akin to the original language model head, \ours does not add complexity to the serving system design and is friendly to distributed settings. We will discuss the training recipe for \ours heads in Section~\ref{sec:training_recipe}.

\subsubsection{Tree Attention}
\label{sec:tree_attention}
Through \ours heads, we obtain probability predictions for the subsequent $K+1$ tokens. These predictions enable us to create length-$K+1$ continuations as candidates. While the speculative decoding studies~\citep{leviathan2022fast,chen2023accelerating} suggest sampling a single continuation as the candidate, leveraging multiple candidates during decoding can enhance the expected acceptance length within a decoding step. Nevertheless, more candidates can also raise computational demands. To strike a balance, we employ a tree-structured attention mechanism to process multiple candidates concurrently.
\begin{figure}[ht]
    \centering
    \includegraphics[width=0.45\textwidth]{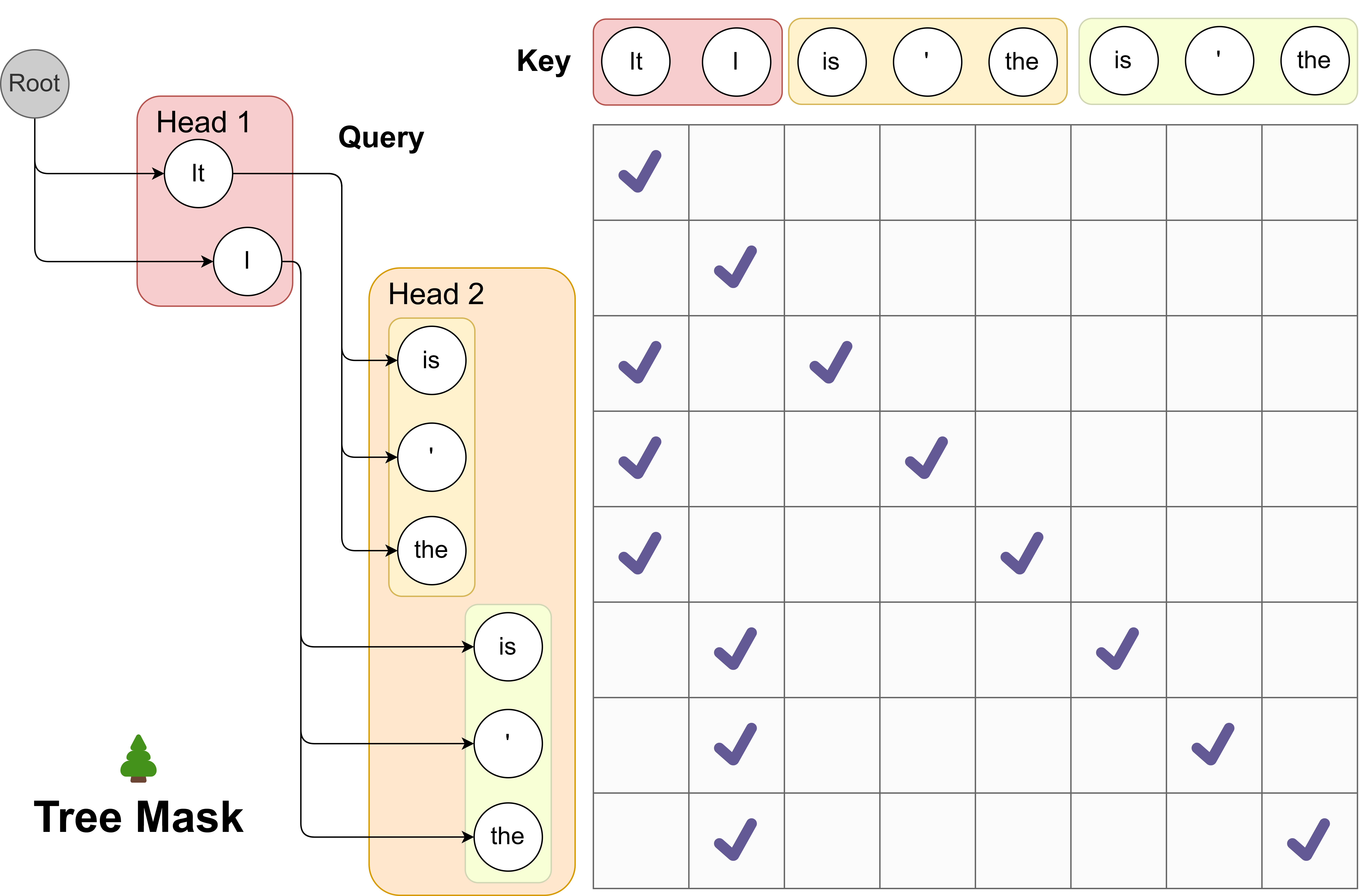}
    \caption{
    We demonstrates the use of tree attention to process multiple candidates concurrently. As exemplified, the top-2 predictions from the first \ours head and the top-3 from the second result in a total of $2\times3=6$ candidates. Each of these candidates corresponds to a distinct branch within the tree structure. To guarantee that each token only accesses its predecessors, we devise an attention mask that exclusively permits attention flow from the current token back to its antecedent tokens. The positional indices for positional encoding are adjusted in line with this structure.}
    \label{fig:tree_attention}
\end{figure}
This attention mechanism diverges from the traditional causal attention paradigm. Within this framework, only tokens from the same continuation are regarded as historical data. Drawing inspiration from the concept of embedding graph structures into attention as proposed in the graph neural network domain~\citep{ying2021transformers}, we incorporate the tree structure into our attention mask, visualized in Figure~\ref{fig:tree_attention}. Remarkably, similar ideas have also been explored in independent works like \citet{miao2023specinfer,spector2023accelerating}, where they follow a bottom-up approach and construct the tree by merging multiple candidates generated by a draft model. In our method, we instead take a top-down approach to build the tree thanks to the structure of candidates generated by \ours heads. For a given $k$-th head, its top-$s_k$ predictions serve as the basis for candidate formation, where $s_k$ is a designated hyperparameter. These candidates are established by determining the Cartesian product of the top-$s_k$ predictions from each head. For instance, in Figure~\ref{fig:tree_attention}, with $s_1=2$ and $s_2=3$, each first head prediction can be succeeded by any prediction from the second head. This leads to a tree structure where $s_k$ branches exist at the $k$-th level (considering a virtual root as the $0$-level, in practice, this $0$-level is for the prediction of the language model head of the original model, which can be sampled independently). Within this tree, only a token's predecessors are seen as historical context, and our attention mask ensures that the attention is only applied on a token's predecessors. By employing this mask and properly setting the positional indices for positional encoding, we can process numerous candidates simultaneously without the need to expand the batch size. The cumulative number of new tokens is calculated as $\sum_{k=1}^K \prod_{i=1}^k s_i$.

In this section, we demonstrate the most simple and regular way to construct the tree structure by taking the Cartesian product. However, it is possible to construct the tree structure in a more sophisticated way and exploit the unbalanced accuracy of different top predictions of different heads. We will discuss this in Section~\ref{sec:optimized_tree_construction}.
\subsection{Training Strategies}
\label{sec:training_recipe}
At the most basic level, we can train \ours heads by freezing the backbone model and fine-tuning \ours heads. However, training the backbone in conjunction with the \ours heads can significantly enhance the accuracy of the \ours heads. Depending on the computational resources and the specific reqirements of the use case, we propose two levels of training strategies for \ours heads.

In this section, we assume the availability of a training dataset that aligns with the target model’s output distribution. This could be the dataset used for Supervised Fine-Tuning (SFT) of the target model. We will discuss eliminating the need for such a dataset using a self-distillation approach in Section~\ref{sec:self_distillation}.
\subsubsection{\ours-1: Frozen Backbone}
\label{sec:frozen_backbone}
To train \ours heads with a frozen backbone model, we can use the cross-entropy loss between the prediction of \ours heads and the ground truth. Specifically, given the ground truth token $y_{t+k+1}$ at position $t+k+1$, the loss for the $k$-th head is $\mathcal{L}_k = -\log p_t^{(k)}(y_{t+k+1})$ where $p_t^{(k)}(y)$ denotes the probability of token $y$ predicted by the $k$-th head. We also observe that $\mathcal{L}_k$ is larger when $k$ is larger, which is reasonable since the prediction of the $k$-th head is more uncertain when $k$ is larger. Therefore, we can add a weight $\lambda_k$ to $\mathcal{L}_k$ to balance the loss of different heads. And the total \ours loss is:
\begin{align}
    \label{eq:loss_medusa_1}
    \mathcal{L}_{\text{\ours-1}} = \sum_{k=1}^K -\lambda_k\log p_t^{(k)}(y_{t+k+1}).
\end{align}

In practice, we set $\lambda_k$ as the $k$-th power of a constant like $0.8$. Since we only use the backbone model for providing the hidden states, we can use a quantized version of the backbone model to reduce the memory consumption. This introduces a more democratized way to accelerate LLM inference, as with the quantization, \ours can be trained for a large model on a single consumer GPU similar to QLoRA~\citep{dettmers2023qlora}. The training only takes a few hours (e.g., 5 hours for \ours-1 on Vicuna 7B model with a single NVIDIA A100 PCIE GPU to train on 60k ShareGPT samples).
\subsubsection{\ours-2: Joint Training}
\label{sec:joint_training}
To further improve the accuracy of \ours heads, we can train \ours heads together with the backbone model. However, this requires a special training recipe to preserve the backbone model's next-token prediction capability and output quality. To achieve this, we propose three strategies:
\begin{itemize}
    \item \textbf{Combined loss}: To keep the backbone model's next-token prediction capability, we need to add the cross-entropy loss of the backbone model $\mathcal{L}_{\text{LM}}=-\log p_t^{(0)}(y_{t+1})$ to the \ours loss. We also add a weight $\lambda_0$ to balance the loss of the backbone model and the \ours heads. Therefore, the total loss is:
    \begin{align}
        \label{eq:loss_medusa_2}
        \mathcal{L}_{\text{\ours-2}} = \mathcal{L}_{\text{LM}} + \lambda_0\mathcal{L}_{\text{\ours-1}}.
    \end{align}
    \item \textbf{Differential learning rates}: Since the backbone model is already well-trained and the \ours heads need more training, we can use separate learning rates for them to enable faster convergence of \ours heads while preserving the backbone model's capability.
    \item \textbf{Heads warmup}: Noticing that at the beginning of training, the \ours heads have a large loss, which leads to a large gradient and may distort the backbone model's parameters. Following the idea from \citet{kumar2022finetuning}, we can employ a two-stage training process. In the first stage, we only train the \ours heads as \ours-1. In the second stage, we train the backbone model and \ours heads together with a warmup strategy. Specifically, we first train the backbone model for a few epochs, then train the \ours heads together with the backbone model. Besides this simple strategy, we can also use a more sophisticated warmup strategy by gradually increasing the weight $\lambda_0$ of the backbone model's loss. We find both strategies work well in practice.
\end{itemize}
Putting these strategies together, we can train \ours heads together with the backbone model without hurting the backbone model's capability. Moreover, this recipe can be applied together with Supervised Fine-Tuning (SFT), enabling us to get a model with native \ours support.
\subsubsection{How to Select the Number of Heads}
Empirically, we found that five heads are sufficient at most. Therefore, we recommend training with five heads and referring to the strategy described in Section~\ref{sec:optimized_tree_construction} to determine the optimal configuration of the tree attention. With optimized tree attention, sometimes three or four heads may be enough for inference. In this case, we can ignore the redundant heads without overhead.

\subsection{Extensions}
\subsubsection{Typical Acceptance}
\label{sec:typical_acceptance}
In speculative decoding papers~\citep{leviathan2022fast,chen2023accelerating}, authors employ rejection sampling to yield diverse outputs that align with the distribution of the original model. However, subsequent implementations~\citep{gante2023assisted,spector2023accelerating} reveal that this sampling strategy results in diminished efficiency as the sampling temperature increases. Intuitively, this can be comprehended in the extreme instance where the draft model is the same as the original one\textcolor{black}{:} 
Using greedy decoding, all output of the draft model will be accepted, therefore maximizing the efficiency. 
Conversely, rejection sampling introduces extra overhead, as the draft model and the original model are sampled independently. Even if their distributions align perfectly, the output of the draft model may still be rejected.

However, in real-world scenarios, sampling from language models is often employed to generate diverse responses, and the temperature parameter is used merely to modulate the ``creativity'' of the response. Therefore, higher temperatures should result in more opportunities for the original model to accept the draft model's output. We ascertain that it is typically unnecessary to match the distribution of the original model. Thus, we propose employing a \emph{typical acceptance} scheme to select plausible candidates rather than using rejection sampling. This approach draws inspiration from truncation sampling studies~\citep{hewitt2022truncation} (refer to \textcolor{black}{Appendix}~\ref{sec:related_work} for an in-depth explanation). Our objective is to choose candidates that are \emph{typical}, meaning they are not exceedingly improbable to be produced by the original model. We use the prediction probability from the \emph{original model} as a natural gauge for this and establish a threshold based on the prediction distribution to determine acceptance. Specifically, given $x_1, x_2, \cdots, x_n$ as context, when evaluating the candidate sequence \textcolor{black}{$(x_{n+1}, x_{n+2}, \cdots, x_{n+K+1})$ }(composed by top predictions of the original language model head and \ours heads), we consider the condition
\begin{align*}
p_{\text{original}}(x_{n+k}|x_1, x_2, \cdots, x_{n+k-1}) > \\\min\rbr{\epsilon, \delta\exp\rbr{-H(p_{\text{original}}(\cdot|x_1, x_2, \cdots, x_{n+k-1}))}},
\end{align*}
where $H(\cdot)$ denotes the entropy function, and $\epsilon, \delta$ are \textcolor{black}{the hard threshold and the entropy-dependent
threshold respectively}. This criterion is adapted from \citet{hewitt2022truncation} and rests on two observations: (1) tokens with relatively high probability are meaningful, and (2) when the distribution's entropy is high, various continuations may be deemed reasonable. During decoding, every candidate is evaluated using this criterion, and a \emph{prefix} of the candidate is accepted if it satisfies the condition. To guarantee the generation of at least one token at each step, we apply \emph{greedy decoding} for the first token and \emph{unconditionally} accept it while employing typical acceptance for subsequent tokens. The final prediction for the current step is determined by the \emph{longest accepted prefix} among all candidates.

Examining this scheme leads to several insights. Firstly, when the temperature is set to $0$, it reverts to greedy decoding, as only the most probable token possesses non-zero probability. As the temperature surpasses $0$, the outcome of greedy decoding will consistently be accepted with appropriate $\epsilon, \delta$, since those tokens have the maximum probability, yielding maximal speedup. Likewise, in general scenarios, an increased temperature will correspondingly result in longer accepted sequences, as corroborated by our experimental findings.

Empirically, we verify that typical acceptance can achieve a better speedup while maintaining a similar \textcolor{black}{generation quality} as shown in Figure~\ref{fig:threshold_ablation}.
\subsubsection{Self-Distillation}
\label{sec:self_distillation}
In Section~\ref{sec:training_recipe}, we assume the existence of a training dataset that matches the target model's output distribution. However, this is not always the case. For example, the model owners may only release the model without the training data, or the model may have gone through a Reinforcement Learning with Human Feedback (RLHF) procedure, which makes the output distribution of the model different from the training dataset. To tackle this issue, we propose an automated self-distillation pipeline to use the model itself to generate the training dataset for \ours heads, which matches the output distribution of the model.

The dataset generation process is straightforward. We first take a public seed dataset from a domain similar to the target model; for example, using the ShareGPT~\citep{sharegpt2023} dataset for chat models. Then, we simply take the prompts from the dataset and ask the model to reply to the prompts. In order to obtain multi-turn conversation samples, we can sequentially feed the prompts from the seed dataset to the model. Or, for models like Zephyr 7B~\citep{tunstall2023zephyr}, which are trained on both roles of the conversation, they have the ability to self-talk, and we can simply feed the first prompt and let the model generate multiple rounds of conversation.

For \ours-1, this dataset is sufficient for training \ours heads. However, for \ours-2, we observe that solely using this dataset for training the backbone and \ours heads usually leads to a lower generation quality. In fact, even without training \ours heads, training the backbone model with this dataset will lead to performance degradation. This suggests that we also need to use the original model's probability prediction instead of using the ground truth token as the label for the backbone model, similar to classic knowledge distillation works~\citep{kim2016sequencelevel}. Concretely, the loss for the backbone model is:
\begin{align*}
    \mathcal{L}_{\text{LM-distill}} = KL(p_{\text{original},t}^{(0)}||p_t^{(0)}),
\end{align*}
where $p_{\text{original},t}^{(0)}$ denotes the probability distribution of the original model's prediction at position $t$.

However, naively, to obtain the original model's probability prediction, we need to maintain two models during training, increasing the memory requirements. To further alleviate this issue, we propose a simple yet effective way to exploit the self-distillation setup. We can use a parameter-efficient adapter like LoRA~\citep{hu2021lora} for fine-tuning the backbone model. In this way, the original model is simply the model with the adapter turned off. Therefore, the distillation does not require additional memory consumption. Together, this self-distillation pipeline can be used to train \ours-2 without hurting the backbone model's capability and introduce almost no additional memory consumption. Lastly, one tip about using self-distillation is that it is preferable to use LoRA without quantization in this case, otherwise, the teacher model will be the quantized model, which may lead to a lower generation quality.

\subsubsection{Searching for the Optimized Tree Construction}
\label{sec:optimized_tree_construction}
In Section~\ref{sec:tree_attention}, we present the simplest way to construct the tree structure by taking the Cartesian product. However, with a fixed budget for the number of total nodes in the tree, a regular tree structure may not be the best choice. Intuitively, those candidates composed of the top predictions of different heads may have different accuracies. Therefore, we can leverage an estimation of the accuracy to construct the tree structure.

Specifically, we can use a calibration dataset and calculate the accuracies of the top predictions of different heads. Let $a_k^{(i)}$ denote the accuracy of the $i$-th top prediction of the $k$-th head\footnote{Here, the accuracy is defined for the single top $i$-th token, i.e., this accuracy is equal to top-$i$ accuracy minus top-$(i-1)$ accuracy.}. Assuming the accuracies are independent, we can estimate the accuracy of a candidate sequence composed by the top $\sbr{i_1, i_2, \cdots, i_k}$ predictions of different heads as $\prod_{j=1}^k a_j^{(i_j)}$. Let $I$ denote the set of all possible combinations of $\sbr{i_1, i_2, \cdots, i_k}$ and each element of $I$ can be mapped to a node of the tree (not only leaf nodes but all nodes are included). Then, the expectation of the acceptance length of a candidate sequence is:
\begin{align*}
    \sum_{\sbr{i_1, i_2, \cdots, i_k}\in I}\prod_{j=1}^k a_j^{(i_j)}.
\end{align*}
Thinking about building a tree by adding nodes one by one, the contribution of a new node to the expectation is exactly the accuracy associated with the node. Therefore, we can greedily add nodes to the tree by choosing the node that is connected to the current tree and has the highest accuracy. This process can be repeated until the total number of nodes reaches the desired number. In this way, we can construct a tree that maximizes the expectation of the acceptance length. Further details can be found in Appendix~\ref{appendix:sparse_tree}.

\begin{figure*}[h]
     \centering
     \begin{subfigure}[b]{0.45\textwidth}
         \centering
         \includegraphics[width=\textwidth]{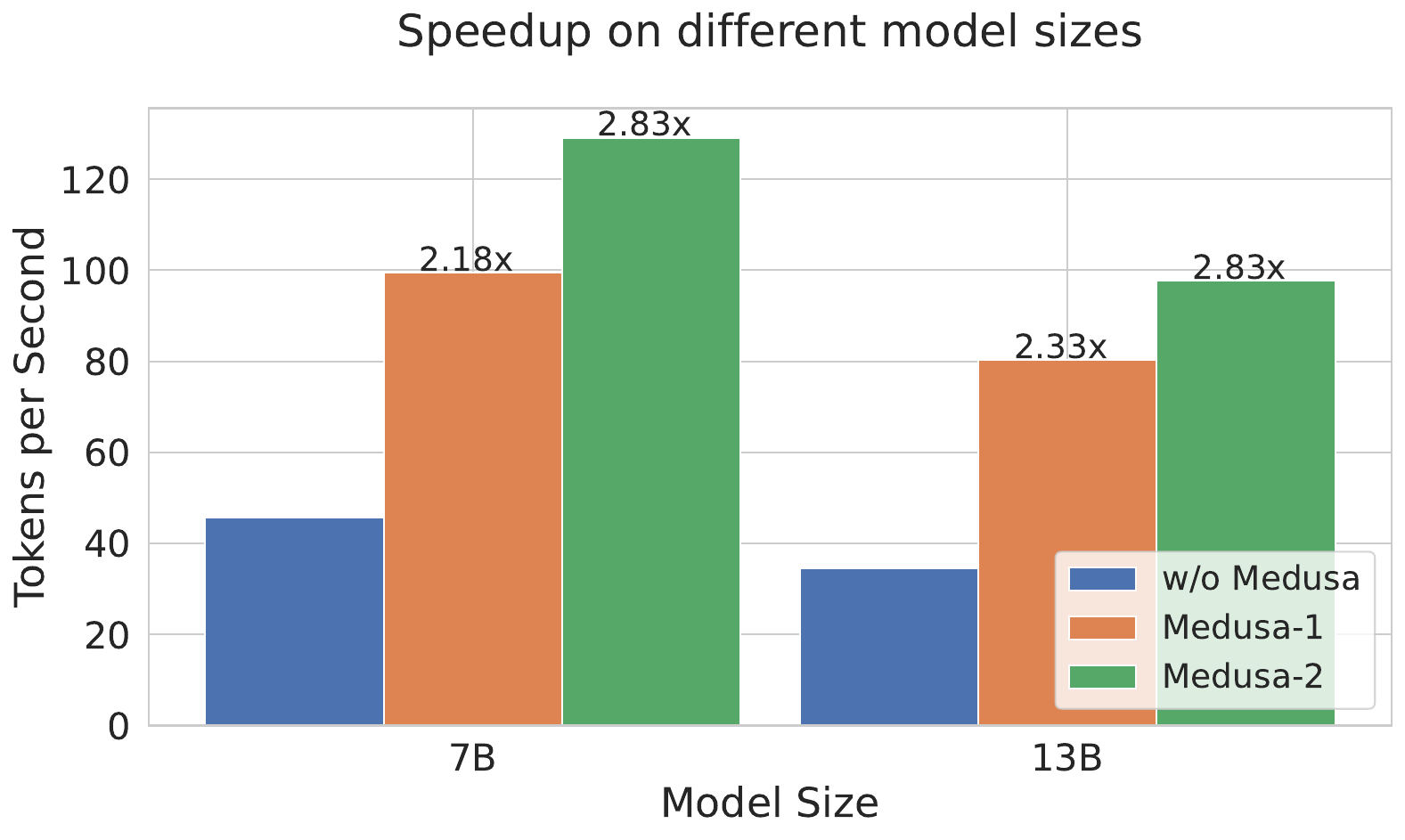}
         \caption{}
         \label{fig:speedup_model_sizes}
     \end{subfigure}
     \begin{subfigure}[b]{0.45\textwidth}
         \centering
         \includegraphics[width=\textwidth]{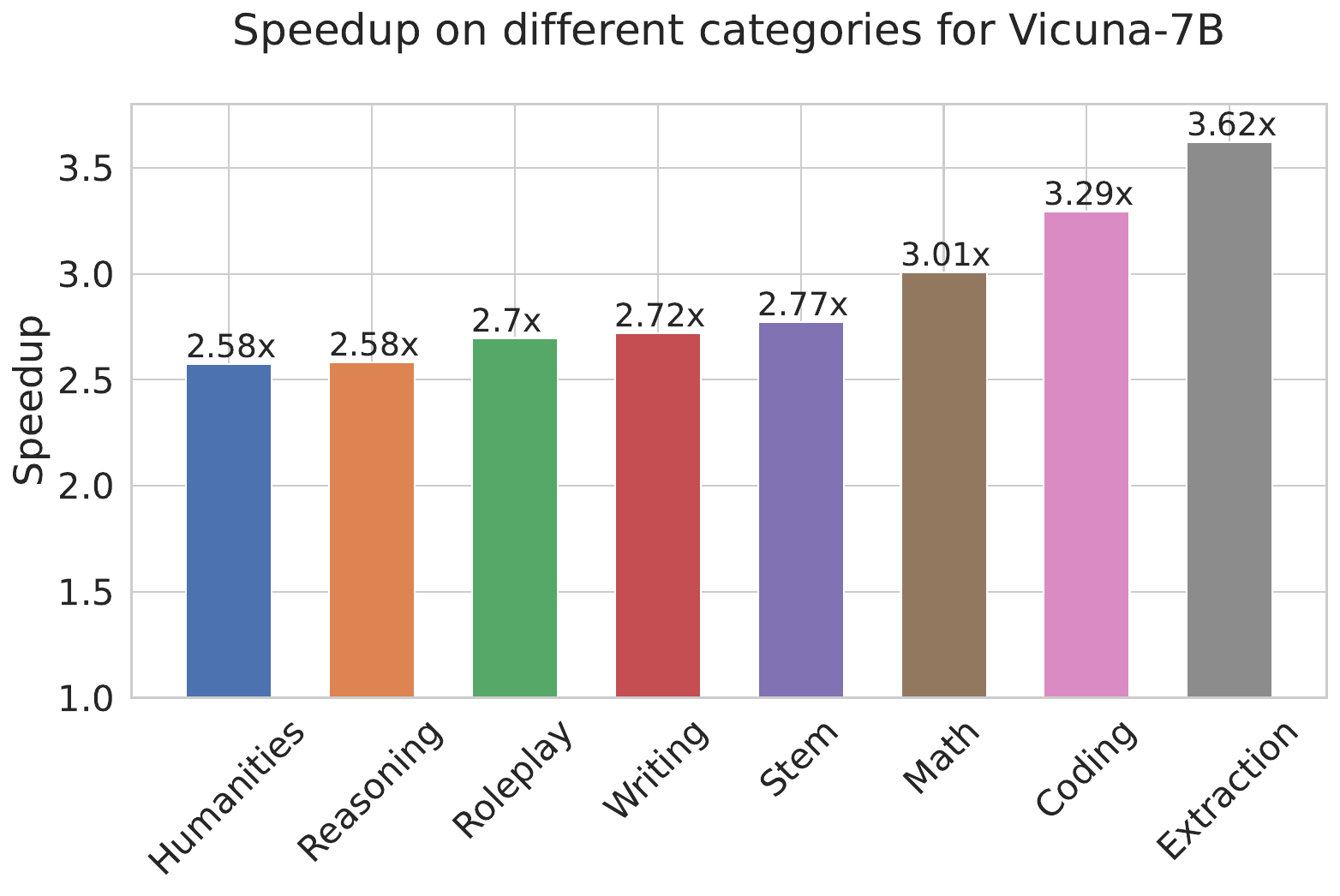}
         \caption{}
         \label{fig:speedup_per_class}
     \end{subfigure}
        \caption{Left: Speed comparison of baseline, \ours-1 and \ours-2 on Vicuna-7B/13B. \ours-1 achieves more than 2$\times$ wall-time speedup compared to the baseline implementation while \ours-2 further improves the speedup by a significant margin. Right: Detailed speedup performance of Vicuna-7B with \textcolor{black}{\ours-2} on 8 categories from MT-Bench.}
        \label{fig:ablation_tree}
\end{figure*}

\section{Experiments}
In this section, we present experiments to demonstrate the effectiveness of \ours under different settings. First, we evaluate \ours on the Vicuna-7B and 13B models~\citep{vicuna2023} to show the performance of \ours-1 and \ours-2. 
\textcolor{black}{Then, we assess our method using the Vicuna-33B and Zephyr-7B models to demonstrate self-distillation's viability in scenarios where direct access to the fine-tuning recipe is unavailable, as with Vicuna-33B, and in models like Zephyr-7B that employ Reinforcement Learning from Human Feedback (RLHF). The evaluation is conducted on MT-Bench~\citep{zheng2023judging}, a multi-turn, conversational-format benchmark.}
Detailed settings can be found in Appendix~\ref{appendix:experiment_settings}.

\subsection{Case Study: \ours-1 v.s. \ours-2 on Vicuna 7B and 13B}

\textbf{Experimental Setup.}
We use the Vicuna model class~\citep{vicuna2023}, which encompasses chat models of varying sizes (7B, 13B, 33B) that are fine-tuned from the Llama model~\citep{touvron2023llama}. Among them, the 7B and 13B models are trained on the ShareGPT~\citep{sharegpt2023} dataset, while the 33B model is an experimental model and is trained on a private dataset. 
In this section, we use the ShareGPT dataset to train the \ours heads on the 7B and 13B models for $2$ epochs. We use the v1.5 version of Vicuna models, which are fine-tuned from Llama-2 models with sequence length 4096.

\textbf{Results.}
We collect the results and show them in Fig.~\ref{fig:ablation_tree}. The baseline is the 
\textcolor{black}{default}
Huggingface implementation. In Fig.~\ref{fig:speedup_model_sizes}, we can see that for the 7B models, \ours-1 and \ours-2 configurations lead to a significant increase in speed, measuring in tokens processed per second. \ours-1 shows a 2.18$\times$ speedup, while \ours-2 further improves this to a 2.83$\times$.
When applied to the larger 13B model, \ours-1 results in a 2.33$\times$ speed increase, while \ours-2 maintains a similar performance gain of 2.83$\times$ over the baseline.
We also plot the speedup per category for \ours-2 Vicuna-7B model. 
We observe that the coding category benefits from a 3.29$\times$ speedup, suggesting that \ours is particularly effective for tasks in this domain. This points to a significant potential for optimizing coding LLMs, which are widely used in software development and other programming-related tasks. 
The ``Extraction'' category shows the highest speedup at 3.62$\times$, indicating that this task is highly optimized by the \ours.
Overall, the results suggest that the \ours significantly enhances inference speed across different model sizes and tasks.

\subsection{Case Study: Training with Self-Distillation on Vicuna-33B and Zephyr-7B}
\textbf{Experimental Setup.}
In this case study, we focus on the cases where self-distillation is needed. We use the Vicuna-33B model~\citep{vicuna2023} and the Zephyr-7B model~\citep{tunstall2023zephyr} as examples. Following the procedure described in Section~\ref{sec:self_distillation}, we first generate the datasets with some seed prompts. We use ShareGPT~\citep{sharegpt2023} and UltraChat~\citep{ding2023enhancing} as the seed datasets and collect a dataset at about $100k$ samples for both cases. Interestingly, we find that the Zephyr model can continue to generate multiple rounds of conversation with a single prompt, which makes it easy to collect a large dataset. 
For Vicuna-33B, we generate the multi-turn conversations by iteratively feeding the prompts from each multi-turn seed conversation \textcolor{black}{using random sampling with temperature 0.3}. Both models are trained with sequence length $2048$ and batch size $128$. 
\begin{table}[h]
\centering
\scriptsize
\begin{tabular}{@{}lllll@{}}
\toprule
Model Name & Vicuna-7B & Zephyr-7B & Vicuna-13B & Vicuna-33B \\ \midrule
Acc. rate   & 3.47      & 3.14      & 3.51       & 3.01       \\ 
Overhead   & 1.22      & 1.18      & 1.23       & 1.27       \\ 
Quality    & 6.18 (+0.01) & 7.25 (-0.07) & 6.43 (-0.14) & 7.18 (+0.05) \\ \midrule
$S_{\textnormal{SpecDecoding}}$ & 1.47 & - & 1.56 & 1.60 \\
$S_\ours$ & 2.83 & 2.66 & 2.83 & 2.35\\\bottomrule
\end{tabular}
\caption{Comparison of various \ours-2 models. 
\textcolor{black}{The first section reports the details of \ours-2, including accelerate rate, overhead, and quality that denoted the average scores on the MT-Bench \textcolor{black}{compared to the original models}. The second section lists the speedup ($S$) of SpecDecoding and \ours, respectively.}
}

\label{tab:all_model_results}
\end{table}

\begin{figure*}[ht]
     \centering
     \begin{subfigure}[b]{0.45\textwidth}
         \centering
         \includegraphics[width=\textwidth]{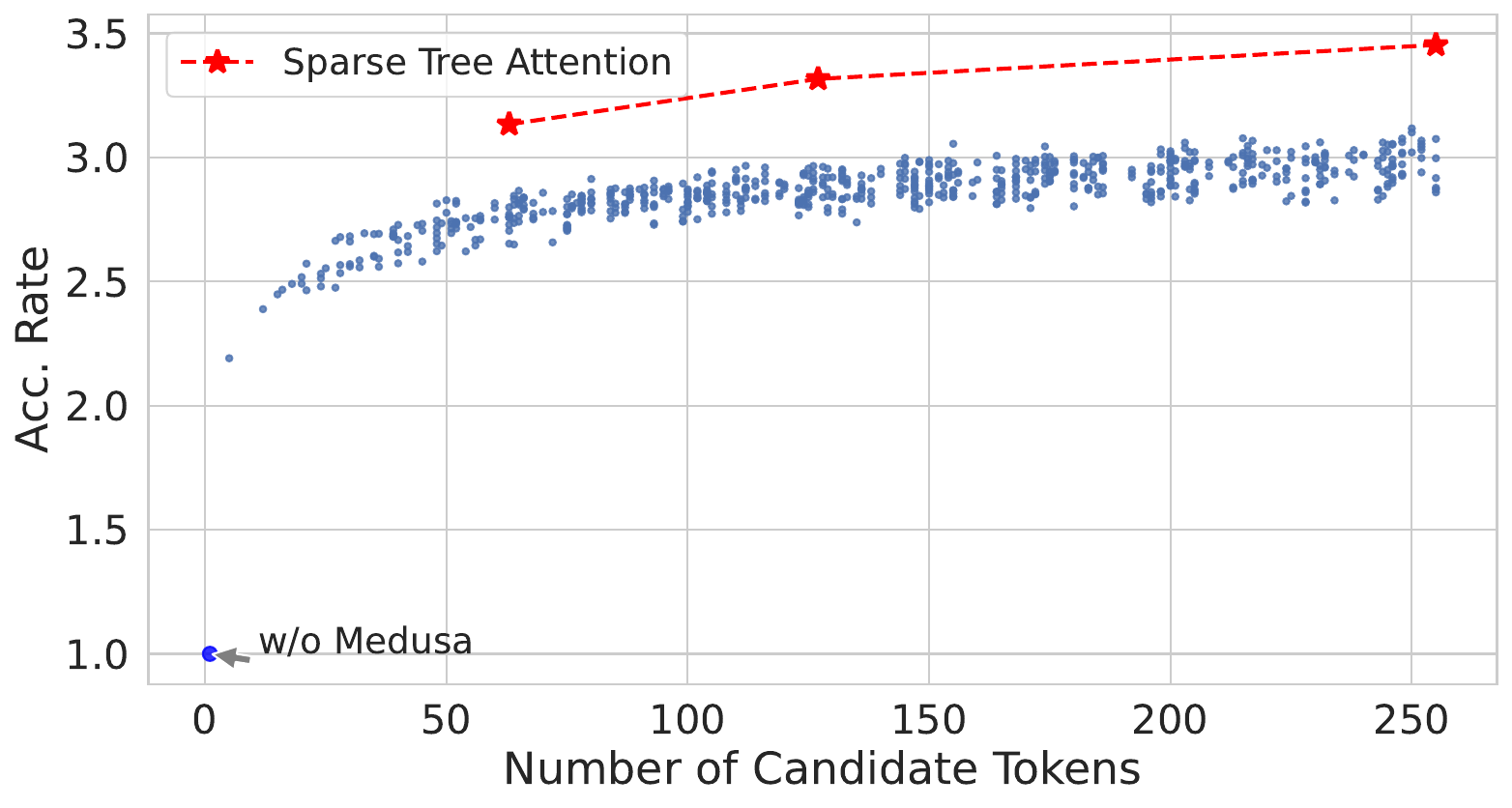}
         \caption{}
         \label{fig:sparse_acc}
     \end{subfigure}
     \begin{subfigure}[b]{0.45\textwidth}
         \centering
         \includegraphics[width=\textwidth]{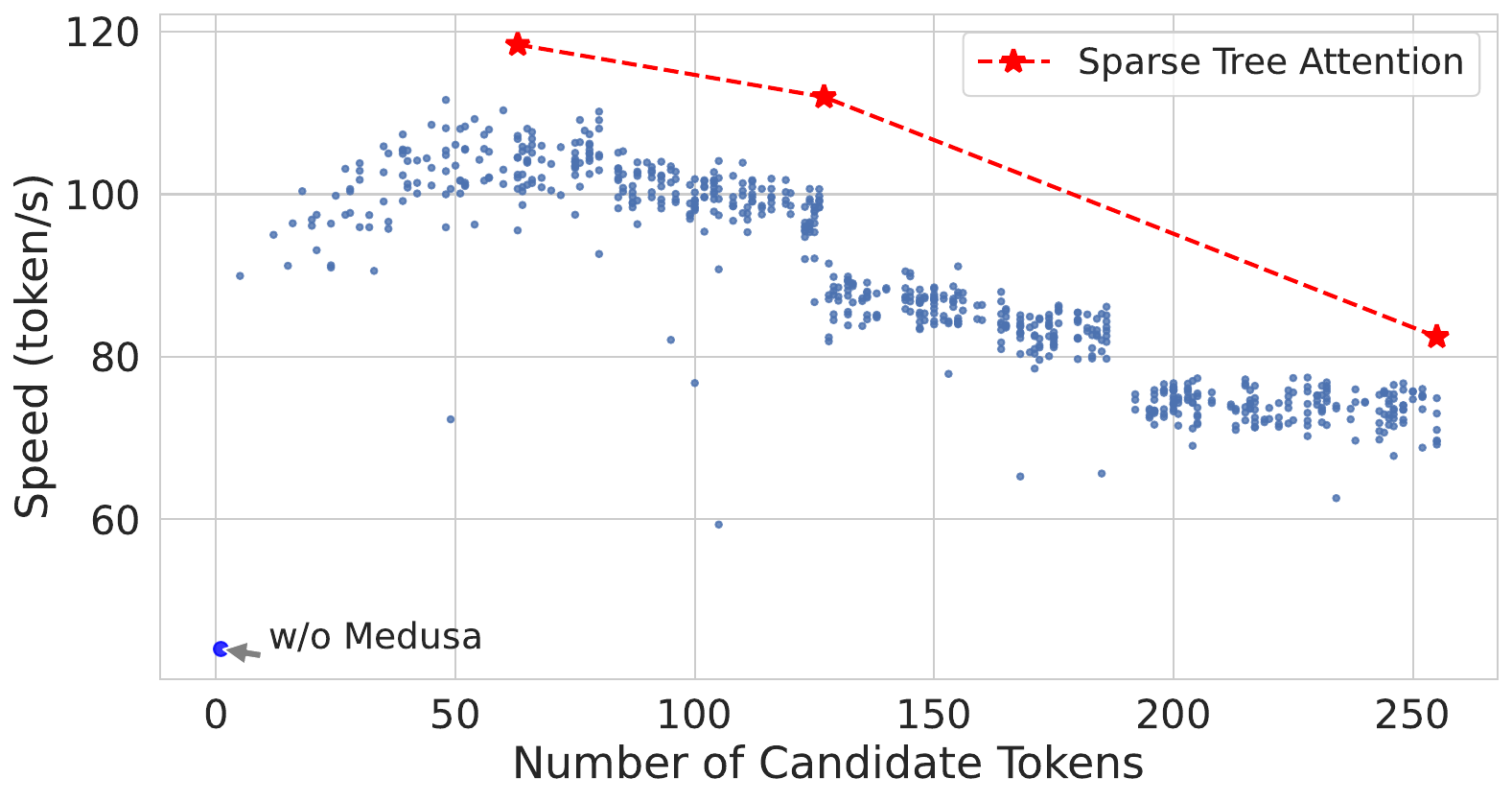}
         \caption{}
         \label{fig:sparse_speed}
     \end{subfigure}
\vspace{-4mm}
        \caption{Effectiveness of numbers of candidate tokens for decoding introduced by trees (default number of candidate token for decoding is 1 when using KV cache). Left: The acceleration rate for randomly sampled dense tree settings (blue dots) and optimized sparse tree settings (red stars). Right: The speed (tokens/s) for both settings. The trend lines indicate that while the \textcolor{black}{acceleration} rate remains relatively stable for sparse trees, there is a notable decrease in speed as the candidate tokens increases.}
        \label{fig:sparse_tree_ablation}
\end{figure*}

\textbf{Results.}
Table~\ref{tab:all_model_results} complements these findings by comparing various \ours-2 models in terms of their acceleration rate, overhead, and quality on MT-Bench \textcolor{black}{with GPT-4 acting as the evaluator to assign performance scores ranging from 0 to 10. We report the quality differences of \ours compared to the original model. 
} 
Notably, while the \ours-2 Vicuna-33B model shows a lower acceleration rate, it maintains a comparable quality. We hypothesize that this is due to a mismatch between the hidden training dataset and the dataset we used for self-distillation. \textcolor{black}{Hence, the model's generation quality can be well aligned by self-distillation while \ours heads learn distribution from the self-distillation that potentially shifts from the training set.}
In our study, we also applied speculative decoding~\citep{chen2023accelerating,leviathan2022fast} to the Vicuna lineup using open-source draft models (details can be found
in Appendix~\ref{appendix:spec}).

These results underscore the complex interplay between speed and performance when scaling up model sizes and applying self-distillation techniques. The findings also highlight the potential of the \ours-2 configuration to boost efficiency in processing while carefully preserving the quality of the model's outputs, suggesting a promising direction for co-optimizing LLMs with \ours heads.

\subsection{Ablation Study}
\subsubsection{Configuration of Tree Attention}~\label{section:config of tree}
The study of tree attention is conducted on the writing and roleplay categories from the MT-Bench dataset using \ours-2 Vicuna-7B. We target to depict tree attention's motivation and its performance. 

Fig.~\ref{fig:sparse_acc} compares the acceleration rate of randomly sampled dense tree configurations (Section.~\ref{sec:tree_attention}, depicted by blue dots) against optimized sparse tree settings (Section.~\ref{sec:optimized_tree_construction}, shown with red stars). The sparse tree configuration with 64 nodes shows a better acceleration rate than the dense tree settings with 256 nodes.
The decline in speed in Fig.~\ref{fig:sparse_speed} is attributed to the increased overhead introduced by the \textcolor{black}{compute-bound. While a more complex tree can improve acceleration, it does so at the cost of speed due to intensive matrix multiplications for linear layers and self-attention. The acceleration rate increase follows a logarithmic trend and slows down when the tree size grows as shown in Fig.~\ref{fig:sparse_acc}. However, the initial gains are substantial, allowing Medusa to achieve significant speedups. If the acceleration increase is less than the overhead, it will slow down overall performance.} For detailed study, please refer to Appendix~\ref{sec:roofline}.

\begin{figure}[H]
    \centering
    \includegraphics[width=0.45\textwidth]{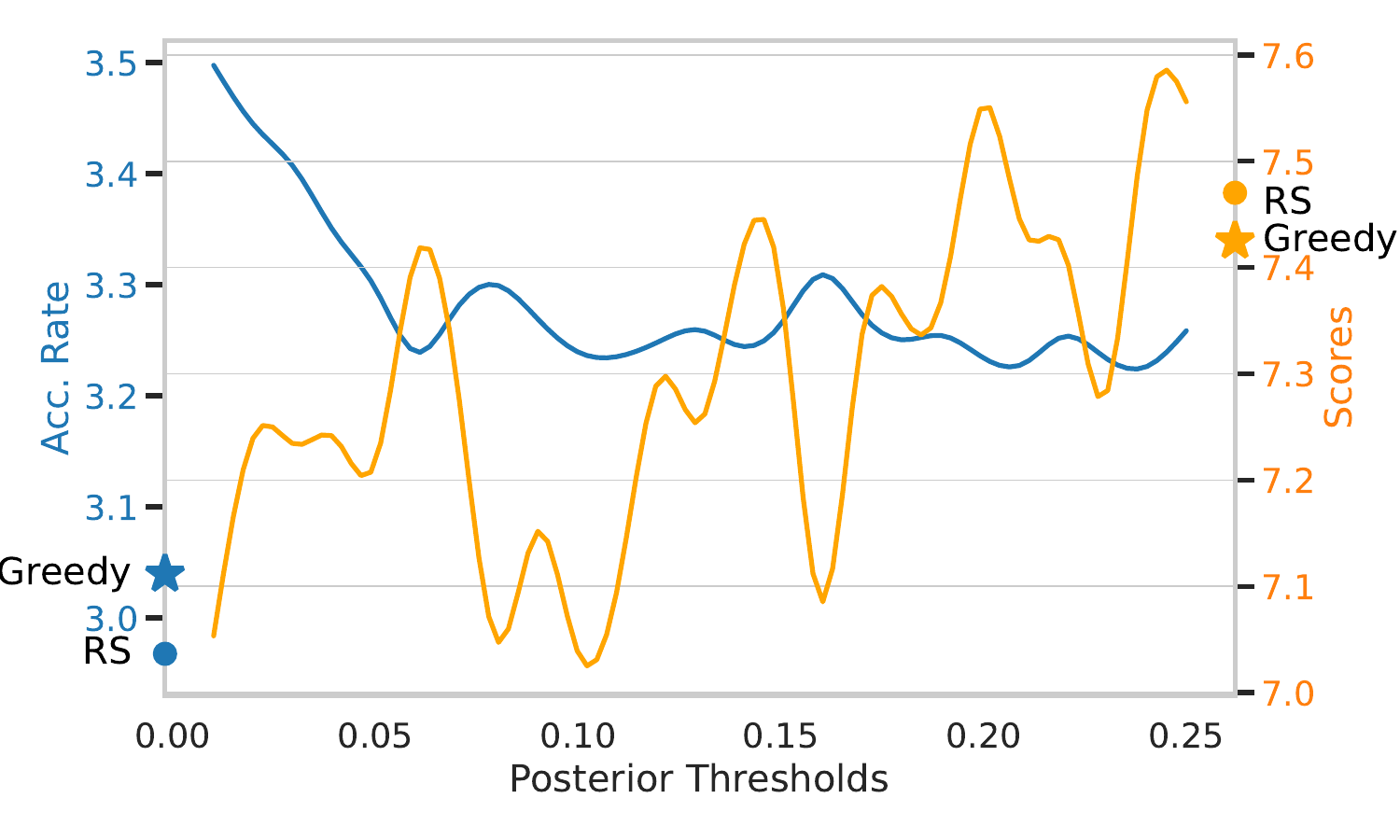}
    \caption{ Performance comparison of \ours using proposed typical sampling. The model is fully fine-tuned from Vicuna-7B. The plot illustrates the acceleration rate and average scores on the writing and roleplay (MT-Bench) with a fixed temperature of 0.7 for 3 different settings: \textcolor{black}{greedy sampling and random sampling (RS) plotted as the star and the dot, and typical sampling curves under different thresholds.}}
    \label{fig:threshold_ablation}
\end{figure}

\subsubsection{Thresholds of Typical Acceptance}
The thresholds of typical acceptance are studied on the writing and roleplay categories from the MT-Bench dataset~\citep{zheng2023judging} using \ours-2 Vicuna 7B. 
Utilizing the Vicuna 7B model, we aligned our methodology with the approach delineated by
~\cite{hewitt2022truncation} setting the $\alpha = \sqrt{\epsilon}$. 
Fig.~\ref{fig:threshold_ablation} presents a comparative analysis of our model's performance across various sampling settings.
These settings range from a threshold $\epsilon$ starting at 0.01 and incrementally increasing to 0.25 in steps of 0.01. Our observations indicate a discernible trade-off: as $\epsilon$ increases, there is an elevation in quality at the expense of a reduced acceleration rate. Furthermore, for tasks demanding creativity, it is noted that the default random sampling surpasses greedy sampling in performance, and the proposed typical sampling is comparable with random sampling when $\epsilon$ increases.

\begin{table}[h!]
\centering
\scriptsize
\begin{tabular}{@{}lcccccc@{}}
\toprule
               & Baseline & Direct Fine-tuning &  \ours-1 & \ours-2 \\ \midrule
Quality        & 6.17        & 5.925            & 6.23        & 6.18        \\ 
Speedup       & N/A                & N/A            & 2.18        & 2.83        \\
\bottomrule
\end{tabular}

\caption{Comparison of Different Settings of Vicuna-7B. Quality is obtained by evaluating models on MT-Bench \textcolor{black}{using GPT-4 as the judge (higher the better)}.}
\label{tab:7b settings}
\end{table}

\subsubsection{Effectiveness of Two-stage Fine-tuning}

\textcolor{black}{
Table~\ref{tab:7b settings} shows the performance differences between various fine-tuning strategies for the Vicuna-7B model. \ours-1, which fine-tunes only the \ours heads, achieves a 2.18x speedup without compromising generation quality. \ours-2, which employs two-stage fine-tuning (Section~\ref{sec:joint_training}), maintains generation quality and provides greater speedup (2.83x) compared to \ours-1. In contrast, direct fine-tuning the model with the \ours heads results in degraded generation quality. 
}
The findings indicate that implementing our \ours-2 for fine-tuning maintains the model's quality and concurrently improves the speedup versus \ours-1.

\begin{table}[h]
    \centering
    \caption{Impact of Techniques on Speedup}
    \begin{tabular}{lc}
    \toprule
    Technique & Speedup \\ \midrule
    Medusa-1 heads without tree attention & $\sim$1.5x \\
    Adding tree attention & $\sim$1.9x \\
    Using optimized tree configuration & $\sim$2.2x \\
    Training heads with Medusa-2 & $\sim$2.8x \\ \bottomrule
    \end{tabular}
    \label{tab:impact_of_tech}
\end{table}

\section{Discussion}
In conclusion, \ours enhances LLM inference speed by 2.3-2.8 times by equipping models with additional predictive decoding heads, allowing for generating multiple tokens simultaneously and bypassing the sequential decoding limitation. Key advantages of \ours include its simplicity, parameter efficiency, and ease of integration into existing systems.
\ours avoids the need for specialized draft models. The typical acceptance scheme removes complications from rejection sampling while providing reasonable outputs. Our approach including two efficient training procedures, ensures high-quality output across various models and prompt types. We summarize the development of each technique and their impact on the speedup in Table~\ref{tab:impact_of_tech}.

In the paper, we focus on the setting with batch size 1 for simplicity. Yet, we want to emphasize that the ideas presented in our paper can be generalized to larger batch-size settings, which are now supported by libraries like TensorRT and Huggingface TGI following our paper. 

\section*{Acknowledgements}
We extend our heartfelt gratitude to several individuals whose contributions were invaluable to this project:
\begin{itemize}
    \item Zhuohan Li, for his invaluable insights on LLM serving. If you haven't already, do check out Zhuohan's vLLM project—it's nothing short of impressive.
    \item Shaojie Bai, for engaging in crucial discussions that helped shape the early phases of this work.
    \item Denny Zhou, for introducing the truncation sampling scheme to Tianle and encouraging Tianle to explore the area of LLM serving.
    \item Yanping Huang, for pointing out the memory-bandwidth-bound challenges associated with LLM serving to Tianle.
    \item Lianmin Zheng, for clarifying the different training recipes used in different sizes of Vicuna models.
\end{itemize}
\textcolor{black}{Jason D. Lee} acknowledges the support of the NSF CCF 2002272, NSF IIS 2107304, and NSF CAREER Award 2144994. \textcolor{black}{Deming Chen acknowledges the support from the AMD Center of Excellence at UIUC.}

\section*{Impact Statement}
The introduction of \ours, an innovative method to improve the inference speed of Large Language Models (LLMs), presents a range of broader implications for society, technology, and ethics. This section explores these implications in detail.

\subsection*{Societal and Technological Implications
}
\begin{itemize}
    \item \textbf{Accessibility and Democratization of AI}: By significantly enhancing the efficiency of LLMs, \ours makes advanced AI technologies more accessible to a wider range of users and organizations. Democratization can spur innovation across various sectors, including education, healthcare, and entertainment, potentially leading to breakthroughs that benefit society at large.
    \item \textbf{Environmental Impact}: The acceleration for LLM inference due to \ours could lead to decreased energy consumption and a smaller carbon footprint. This aligns with the growing need for sustainable AI practices, contributing to environmental conservation efforts.
    \item \textbf{Economic Implications}: The increased efficiency brought about by \ours may lower the cost barrier to deploying state-of-the-art AI models, enabling small and medium-sized enterprises to leverage advanced AI capabilities. This could stimulate economic growth, foster competition, and drive technological innovation.
    \end{itemize}
\subsection*{Ethical Considerations}
\begin{itemize}
    \item \textbf{Bias and Fairness}: While \ours aims to improve LLM efficiency, it inherits the ethical considerations of its backbone models, including issues related to bias and fairness. The method's ability to maintain generation quality necessitates investigation to ensure that the models do not perpetuate or amplify existing biases.

 \item \textbf{Transparency and Accountability}: The complexity of \ours, particularly with its tree-based attention mechanism and multiple decoding heads, may pose challenges in terms of model interpretability. Ensuring transparency in how decisions are made and maintaining accountability for those decisions are crucial for building trust in AI systems.

\item \textbf{Security and Privacy}: The accelerated capabilities of LLMs augmented by \ours could potentially be exploited for malicious purposes, such as generating disinformation at scale or automating cyber-attacks. It is imperative to develop and enforce ethical guidelines and security measures to prevent misuse.

\end{itemize}

\bibliography{icml/medusa_icml}

\begin{thebibliography}{52}
\providecommand{\natexlab}[1]{#1}
\providecommand{\url}[1]{\texttt{#1}}
\expandafter\ifx\csname urlstyle\endcsname\relax
  \providecommand{\doi}[1]{doi: #1}\else
  \providecommand{\doi}{doi: \begingroup \urlstyle{rm}\Url}\fi

\bibitem[Ainslie et~al.(2023)Ainslie, Lee-Thorp, de~Jong, Zemlyanskiy,
  Lebr{\'o}n, and Sanghai]{ainslie2023gqa}
Ainslie, J., Lee-Thorp, J., de~Jong, M., Zemlyanskiy, Y., Lebr{\'o}n, F., and
  Sanghai, S.
\newblock Gqa: Training generalized multi-query transformer models from
  multi-head checkpoints.
\newblock \emph{arXiv preprint arXiv:2305.13245}, 2023.

\bibitem[Axolotl(2023)]{axolotl2023}
Axolotl.
\newblock {Axolotl}.
\newblock \url{https://github.com/OpenAccess-AI-Collective/axolotl}, 2023.

\bibitem[Basu et~al.(2021)Basu, Ramachandran, Keskar, and
  Varshney]{basu2021mirostat}
Basu, S., Ramachandran, G.~S., Keskar, N.~S., and Varshney, L.~R.
\newblock {\{}MIROSTAT{\}}: A {\{}neural{\}} {\{}text{\}} {\{}decoding{\}}
  {\{}algorithm{\}} {\{}that{\}} {\{}directly{\}} {\{}controls{\}}
  {\{}perplexity{\}}.
\newblock In \emph{International Conference on Learning Representations}, 2021.
\newblock URL \url{https://openreview.net/forum?id=W1G1JZEIy5_}.

\bibitem[Brown et~al.(2020)Brown, Mann, Ryder, Subbiah, Kaplan, Dhariwal,
  Neelakantan, Shyam, Sastry, Askell, et~al.]{brown2020language}
Brown, T., Mann, B., Ryder, N., Subbiah, M., Kaplan, J.~D., Dhariwal, P.,
  Neelakantan, A., Shyam, P., Sastry, G., Askell, A., et~al.
\newblock Language models are few-shot learners.
\newblock \emph{Advances in neural information processing systems},
  33:\penalty0 1877--1901, 2020.

\bibitem[Chen et~al.(2023)Chen, Borgeaud, Irving, Lespiau, Sifre, and
  Jumper]{chen2023accelerating}
Chen, C., Borgeaud, S., Irving, G., Lespiau, J.-B., Sifre, L., and Jumper, J.
\newblock Accelerating large language model decoding with speculative sampling.
\newblock February 2023.
\newblock \doi{10.48550/ARXIV.2302.01318}.

\bibitem[Chen(2023)]{chen2023transformer}
Chen, L.
\newblock Dissecting batching effects in gpt inference.
\newblock \url{https://le.qun.ch/en/blog/2023/05/13/transformer-batching/},
  2023.
\newblock Blog.

\bibitem[Chiang et~al.(2023)Chiang, Li, Lin, Sheng, Wu, Zhang, Zheng, Zhuang,
  Zhuang, Gonzalez, Stoica, and Xing]{vicuna2023}
Chiang, W.-L., Li, Z., Lin, Z., Sheng, Y., Wu, Z., Zhang, H., Zheng, L.,
  Zhuang, S., Zhuang, Y., Gonzalez, J.~E., Stoica, I., and Xing, E.~P.
\newblock Vicuna: An open-source chatbot impressing gpt-4 with 90\%* chatgpt
  quality, March 2023.
\newblock URL \url{https://lmsys.org/blog/2023-03-30-vicuna/}.

\bibitem[Chowdhery et~al.(2022)Chowdhery, Narang, Devlin, Bosma, Mishra,
  Roberts, Barham, Chung, Sutton, Gehrmann, et~al.]{chowdhery2022palm}
Chowdhery, A., Narang, S., Devlin, J., Bosma, M., Mishra, G., Roberts, A.,
  Barham, P., Chung, H.~W., Sutton, C., Gehrmann, S., et~al.
\newblock Palm: Scaling language modeling with pathways.
\newblock \emph{arXiv preprint arXiv:2204.02311}, 2022.

\bibitem[Dettmers et~al.(2021)Dettmers, Lewis, Shleifer, and
  Zettlemoyer]{dettmers20218bit}
Dettmers, T., Lewis, M., Shleifer, S., and Zettlemoyer, L.
\newblock 8-bit optimizers via block-wise quantization.
\newblock \emph{International Conference on Learning Representations}, 2021.

\bibitem[Dettmers et~al.(2022)Dettmers, Lewis, Belkada, and
  Zettlemoyer]{dettmers2022llm}
Dettmers, T., Lewis, M., Belkada, Y., and Zettlemoyer, L.
\newblock Llm. int8 (): 8-bit matrix multiplication for transformers at scale.
\newblock \emph{arXiv preprint arXiv:2208.07339}, 2022.

\bibitem[Dettmers et~al.(2023)Dettmers, Pagnoni, Holtzman, and
  Zettlemoyer]{dettmers2023qlora}
Dettmers, T., Pagnoni, A., Holtzman, A., and Zettlemoyer, L.
\newblock Qlora: Efficient finetuning of quantized llms.
\newblock \emph{arXiv preprint arXiv:2305.14314}, 2023.

\bibitem[Ding et~al.(2023)Ding, Chen, Xu, Qin, Zheng, Hu, Liu, Sun, and
  Zhou]{ding2023enhancing}
Ding, N., Chen, Y., Xu, B., Qin, Y., Zheng, Z., Hu, S., Liu, Z., Sun, M., and
  Zhou, B.
\newblock Enhancing chat language models by scaling high-quality instructional
  conversations, 2023.

\bibitem[Dubois et~al.(2023)Dubois, Li, Taori, Zhang, Gulrajani, Ba, Guestrin,
  Liang, and Hashimoto]{dubois2023alpacafarm}
Dubois, Y., Li, X., Taori, R., Zhang, T., Gulrajani, I., Ba, J., Guestrin, C.,
  Liang, P., and Hashimoto, T.~B.
\newblock Alpacafarm: A simulation framework for methods that learn from human
  feedback, 2023.

\bibitem[Elfwing et~al.(2017)Elfwing, Uchibe, and
  Doya]{elfwing2017sigmoidweighted}
Elfwing, S., Uchibe, E., and Doya, K.
\newblock Sigmoid-weighted linear units for neural network function
  approximation in reinforcement learning.
\newblock \emph{Neural Networks}, 2017.
\newblock \doi{10.1016/j.neunet.2017.12.012}.

\bibitem[Fan et~al.(2018)Fan, Lewis, and Dauphin]{fan2018hierarchical}
Fan, A., Lewis, M., and Dauphin, Y.
\newblock Hierarchical neural story generation.
\newblock In \emph{Proceedings of the 56th Annual Meeting of the Association
  for Computational Linguistics (Volume 1: Long Papers)}. Association for
  Computational Linguistics, 2018.
\newblock \doi{10.18653/v1/p18-1082}.

\bibitem[Frantar et~al.(2022)Frantar, Ashkboos, Hoefler, and
  Alistarh]{frantar2022gptq}
Frantar, E., Ashkboos, S., Hoefler, T., and Alistarh, D.
\newblock Gptq: Accurate post-training quantization for generative pre-trained
  transformers.
\newblock \emph{arXiv preprint arXiv:2210.17323}, 2022.

\bibitem[Google(2023)]{google2023palm2}
Google.
\newblock Palm 2 technical report, 2023.
\newblock URL \url{https://ai.google/static/documents/palm2techreport.pdf}.

\bibitem[Hewitt et~al.(2022)Hewitt, Manning, and Liang]{hewitt2022truncation}
Hewitt, J., Manning, C.~D., and Liang, P.
\newblock Truncation sampling as language model desmoothing.
\newblock October 2022.
\newblock \doi{10.48550/ARXIV.2210.15191}.

\bibitem[Hoffmann et~al.(2022)Hoffmann, Borgeaud, Mensch, Buchatskaya, Cai,
  Rutherford, Casas, Hendricks, Welbl, Clark, et~al.]{hoffmann2022training}
Hoffmann, J., Borgeaud, S., Mensch, A., Buchatskaya, E., Cai, T., Rutherford,
  E., Casas, D. d.~L., Hendricks, L.~A., Welbl, J., Clark, A., et~al.
\newblock Training compute-optimal large language models.
\newblock \emph{arXiv preprint arXiv:2203.15556}, 2022.

\bibitem[Holtzman et~al.(2020)Holtzman, Buys, Du, Forbes, and
  Choi]{holtzman2020curious}
Holtzman, A., Buys, J., Du, L., Forbes, M., and Choi, Y.
\newblock The curious case of neural text degeneration.
\newblock In \emph{International Conference on Learning Representations}, 2020.
\newblock URL \url{https://openreview.net/forum?id=rygGQyrFvH}.

\bibitem[Hu et~al.(2021)Hu, Shen, Wallis, Allen-Zhu, Li, Wang, and
  Chen]{hu2021lora}
Hu, E.~J., Shen, Y., Wallis, P., Allen-Zhu, Z., Li, Y., Wang, S., and Chen, W.
\newblock Lora: Low-rank adaptation of large language models.
\newblock \emph{ICLR}, 2021.

\bibitem[{Joao Gante}(2023)]{gante2023assisted}
{Joao Gante}.
\newblock Assisted generation: a new direction toward low-latency text
  generation, 2023.
\newblock URL \url{https://huggingface.co/blog/assisted-generation}.

\bibitem[Kim et~al.(2023)Kim, Hooper, Gholami, Dong, Li, Shen, Mahoney, and
  Keutzer]{kim2023squeezellm}
Kim, S., Hooper, C., Gholami, A., Dong, Z., Li, X., Shen, S., Mahoney, M.~W.,
  and Keutzer, K.
\newblock Squeezellm: Dense-and-sparse quantization.
\newblock \emph{arXiv preprint arXiv:2306.07629}, 2023.

\bibitem[Kim \& Rush(2016)Kim and Rush]{kim2016sequencelevel}
Kim, Y. and Rush, A.~M.
\newblock Sequence-level knowledge distillation.
\newblock \emph{EMNLP}, 2016.

\bibitem[Kumar et~al.(2022)Kumar, Raghunathan, Jones, Ma, and
  Liang]{kumar2022finetuning}
Kumar, A., Raghunathan, A., Jones, R., Ma, T., and Liang, P.
\newblock Fine-tuning can distort pretrained features and underperform
  out-of-distribution.
\newblock \emph{International Conference on Learning Representations}, 2022.

\bibitem[Kwon et~al.(2023)Kwon, Li, Zhuang, Sheng, Zheng, Yu, Gonzalez, Zhang,
  and Stoica]{kwon2023efficient}
Kwon, W., Li, Z., Zhuang, S., Sheng, Y., Zheng, L., Yu, C.~H., Gonzalez, J.~E.,
  Zhang, H., and Stoica, I.
\newblock Efficient memory management for large language model serving with
  pagedattention.
\newblock In \emph{Proceedings of the ACM SIGOPS 29th Symposium on Operating
  Systems Principles}, 2023.

\bibitem[Leviathan et~al.(2022)Leviathan, Kalman, and
  Matias]{leviathan2022fast}
Leviathan, Y., Kalman, M., and Matias, Y.
\newblock Fast inference from transformers via speculative decoding.
\newblock November 2022.
\newblock \doi{10.48550/ARXIV.2211.17192}.

\bibitem[Li et~al.(2023)Li, Zhang, Dubois, Taori, Gulrajani, Guestrin, Liang,
  and Hashimoto]{alpaca_eval}
Li, X., Zhang, T., Dubois, Y., Taori, R., Gulrajani, I., Guestrin, C., Liang,
  P., and Hashimoto, T.~B.
\newblock Alpacaeval: An automatic evaluator of instruction-following models.
\newblock \url{https://github.com/tatsu-lab/alpaca_eval}, 2023.

\bibitem[Lin et~al.(2023)Lin, Tang, Tang, Yang, Dang, and Han]{lin2023awq}
Lin, J., Tang, J., Tang, H., Yang, S., Dang, X., and Han, S.
\newblock Awq: Activation-aware weight quantization for llm compression and
  acceleration.
\newblock \emph{arXiv preprint arXiv:2306.00978}, 2023.

\bibitem[Meister et~al.(2022)Meister, Wiher, Pimentel, and
  Cotterell]{meister2022probability}
Meister, C., Wiher, G., Pimentel, T., and Cotterell, R.
\newblock On the probability-quality paradox in language generation.
\newblock March 2022.
\newblock \doi{10.48550/ARXIV.2203.17217}.

\bibitem[Meister et~al.(2023)Meister, Pimentel, Wiher, and
  Cotterell]{meister2023locally}
Meister, C., Pimentel, T., Wiher, G., and Cotterell, R.
\newblock Locally typical sampling.
\newblock \emph{Transactions of the Association for Computational Linguistics},
  11:\penalty0 102--121, 2023.

\bibitem[Miao et~al.(2023)Miao, Oliaro, Zhang, Cheng, Wang, Wong, Chen, Arfeen,
  Abhyankar, and Jia]{miao2023specinfer}
Miao, X., Oliaro, G., Zhang, Z., Cheng, X., Wang, Z., Wong, R. Y.~Y., Chen, Z.,
  Arfeen, D., Abhyankar, R., and Jia, Z.
\newblock Specinfer: Accelerating generative llm serving with speculative
  inference and token tree verification.
\newblock \emph{arXiv preprint arXiv:2305.09781}, 2023.

\bibitem[NVIDIA()]{nvidia_a100_datasheet}
NVIDIA.
\newblock Nvidia a100 tensor core gpu.

\bibitem[OpenAI(2023)]{openai2023gpt4}
OpenAI.
\newblock Gpt-4 technical report, 2023.

\bibitem[Ouyang et~al.(2022)Ouyang, Wu, Jiang, Almeida, Wainwright, Mishkin,
  Zhang, Agarwal, Slama, Ray, et~al.]{ouyang2022training}
Ouyang, L., Wu, J., Jiang, X., Almeida, D., Wainwright, C.~L., Mishkin, P.,
  Zhang, C., Agarwal, S., Slama, K., Ray, A., et~al.
\newblock Training language models to follow instructions with human feedback.
\newblock \emph{arXiv preprint arXiv:2203.02155}, 2022.

\bibitem[Pan(2023)]{tiny_vicuna_1b}
Pan, J.
\newblock Tiny vicuna 1b.
\newblock \url{https://huggingface.co/Jiayi-Pan/Tiny-Vicuna-1B}, 2023.

\bibitem[Pillutla et~al.(2021)Pillutla, Swayamdipta, Zellers, Thickstun,
  Welleck, Choi, and Harchaoui]{pillutla2021mauve}
Pillutla, K., Swayamdipta, S., Zellers, R., Thickstun, J., Welleck, S., Choi,
  Y., and Harchaoui, Z.
\newblock {MAUVE}: Measuring the gap between neural text and human text using
  divergence frontiers.
\newblock In Beygelzimer, A., Dauphin, Y., Liang, P., and Vaughan, J.~W.
  (eds.), \emph{Advances in Neural Information Processing Systems}, 2021.
\newblock URL \url{https://openreview.net/forum?id=Tqx7nJp7PR}.

\bibitem[Pope et~al.(2022)Pope, Douglas, Chowdhery, Devlin, Bradbury, Levskaya,
  Heek, Xiao, Agrawal, and Dean]{pope2022efficiently}
Pope, R., Douglas, S., Chowdhery, A., Devlin, J., Bradbury, J., Levskaya, A.,
  Heek, J., Xiao, K., Agrawal, S., and Dean, J.
\newblock Efficiently scaling transformer inference.
\newblock November 2022.
\newblock \doi{10.48550/ARXIV.2211.05102}.

\bibitem[ShareGPT(2023)]{sharegpt2023}
ShareGPT.
\newblock {ShareGPT}.
\newblock
  \url{https://huggingface.co/datasets/Aeala/ShareGPT_Vicuna_unfiltered}, 2023.

\bibitem[Shazeer(2019)]{shazeer2019fast}
Shazeer, N.
\newblock Fast transformer decoding: One write-head is all you need.
\newblock \emph{arXiv preprint arXiv:1911.02150}, 2019.

\bibitem[Spector \& Re(2023)Spector and Re]{spector2023accelerating}
Spector, B. and Re, C.
\newblock Accelerating llm inference with staged speculative decoding.
\newblock \emph{arXiv preprint arXiv:2308.04623}, 2023.

\bibitem[Stern et~al.(2018)Stern, Shazeer, and Uszkoreit]{stern2018blockwise}
Stern, M., Shazeer, N.~M., and Uszkoreit, J.
\newblock Blockwise parallel decoding for deep autoregressive models.
\newblock \emph{Neural Information Processing Systems}, 2018.

\bibitem[Touvron et~al.(2023)Touvron, Martin, Stone, Albert, Almahairi, Babaei,
  Bashlykov, Batra, Bhargava, Bhosale, et~al.]{touvron2023llama}
Touvron, H., Martin, L., Stone, K., Albert, P., Almahairi, A., Babaei, Y.,
  Bashlykov, N., Batra, S., Bhargava, P., Bhosale, S., et~al.
\newblock Llama 2: Open foundation and fine-tuned chat models.
\newblock \emph{arXiv preprint arXiv:2307.09288}, 2023.

\bibitem[Tunstall et~al.(2023)Tunstall, Beeching, Lambert, Rajani, Rasul,
  Belkada, Huang, von Werra, Fourrier, Habib, Sarrazin, Sanseviero, Rush, and
  Wolf]{tunstall2023zephyr}
Tunstall, L., Beeching, E., Lambert, N., Rajani, N., Rasul, K., Belkada, Y.,
  Huang, S., von Werra, L., Fourrier, C., Habib, N., Sarrazin, N., Sanseviero,
  O., Rush, A.~M., and Wolf, T.
\newblock Zephyr: Direct distillation of lm alignment, 2023.

\bibitem[Xia et~al.(2023)Xia, Ge, Chen, Wei, and Sui]{xia2023speculative}
Xia, H., Ge, T., Chen, S.-Q., Wei, F., and Sui, Z.
\newblock Speculative decoding: Lossless speedup of autoregressive translation,
  2023.
\newblock URL \url{https://openreview.net/forum?id=H-VlwsYvVi}.

\bibitem[Xiao et~al.(2023{\natexlab{a}})Xiao, Lin, Seznec, Wu, Demouth, and
  Han]{xiao2023smoothquant}
Xiao, G., Lin, J., Seznec, M., Wu, H., Demouth, J., and Han, S.
\newblock Smoothquant: Accurate and efficient post-training quantization for
  large language models.
\newblock In \emph{International Conference on Machine Learning}, pp.\
  38087--38099. PMLR, 2023{\natexlab{a}}.

\bibitem[Xiao et~al.(2023{\natexlab{b}})Xiao, Wu, Guo, Li, Zhang, Qin, and
  Liu]{xiao2023survey}
Xiao, Y., Wu, L., Guo, J., Li, J., Zhang, M., Qin, T., and Liu, T.-y.
\newblock A survey on non-autoregressive generation for neural machine
  translation and beyond.
\newblock \emph{IEEE Transactions on Pattern Analysis and Machine
  Intelligence}, 2023{\natexlab{b}}.

\bibitem[Ying et~al.(2021)Ying, Cai, Luo, Zheng, Ke, He, Shen, and
  Liu]{ying2021transformers}
Ying, C., Cai, T., Luo, S., Zheng, S., Ke, G., He, D., Shen, Y., and Liu, T.-Y.
\newblock Do transformers really perform badly for graph representation?
\newblock \emph{Advances in Neural Information Processing Systems},
  34:\penalty0 28877--28888, 2021.

\bibitem[Zhang et~al.(2024)Zhang, Zeng, Wang, and Lu]{zhang2024tinyllama}
Zhang, P., Zeng, G., Wang, T., and Lu, W.
\newblock Tinyllama: An open-source small language model, 2024.

\bibitem[Zhang et~al.(2022)Zhang, Roller, Goyal, Artetxe, Chen, Chen, Dewan,
  Diab, Li, Lin, et~al.]{zhang2022opt}
Zhang, S., Roller, S., Goyal, N., Artetxe, M., Chen, M., Chen, S., Dewan, C.,
  Diab, M., Li, X., Lin, X.~V., et~al.
\newblock Opt: Open pre-trained transformer language models.
\newblock \emph{arXiv preprint arXiv:2205.01068}, 2022.

\bibitem[Zhang et~al.(2023)Zhang, Sheng, Zhou, Chen, Zheng, Cai, Song, Tian,
  R{\'e}, Barrett, et~al.]{zhang2023h}
Zhang, Z., Sheng, Y., Zhou, T., Chen, T., Zheng, L., Cai, R., Song, Z., Tian,
  Y., R{\'e}, C., Barrett, C., et~al.
\newblock H $ \_2 $ o: Heavy-hitter oracle for efficient generative inference
  of large language models.
\newblock \emph{arXiv preprint arXiv:2306.14048}, 2023.

\bibitem[Zheng et~al.(2023)Zheng, Chiang, Sheng, Zhuang, Wu, Zhuang, Lin, Li,
  Li, Xing, Zhang, Gonzalez, and Stoica]{zheng2023judging}
Zheng, L., Chiang, W.-L., Sheng, Y., Zhuang, S., Wu, Z., Zhuang, Y., Lin, Z.,
  Li, Z., Li, D., Xing, E.~P., Zhang, H., Gonzalez, J.~E., and Stoica, I.
\newblock Judging llm-as-a-judge with mt-bench and chatbot arena, 2023.

\end{thebibliography}
\bibliographystyle{icml2024}

\newpage
\appendix
\onecolumn
\section{Related Work}\label{sec:related_work}
\subsection{LLM Inference Acceleration}
The inefficiency of Large Language Model (LLM) inference is primarily attributed to the memory-bandwidth-bound nature of the auto-regressive decoding process. Several methods have been proposed to alleviate this issue, improving inference latency and throughput. Traditionally, batch inference has been employed as a straightforward method to enhance arithmetic intensity and escape memory-bandwidth-bound limitations. However, with LLMs, both model parameters and the Key-Value (KV) cache consume substantial accelerator memory, hindering the utilization of large batch sizes. Existing methods to tackle this problem can be conceptually divided into two main categories: (1) Reducing memory consumption, thereby minimizing memory transfer overhead and enabling larger batch sizes, and (2) Minimizing the number of decoding steps to decrease latency directly.

\paragraph{Reducing KV Cache.} Methods such as Multi-query attention~\citep{shazeer2019fast} and Grouped-query attention~\citep{ainslie2023gqa} adopt a direct approach to diminish the KV cache. By utilizing fewer key and value heads in the attention modules relative to query heads, these strategies substantially cut the KV's memory consumption, thereby facilitating larger batch sizes and enhanced accelerator utilization~\citep{pope2022efficiently}. Additionally, \citet{zhang2023h} proposes to selectively retain the most critical KV tokens, further reducing the KV cache. From a system perspective, \citet{kwon2023efficient} introduces a paged memory management scheme for reducing fragmentation of the KV cache.

\paragraph{Quantization.} Quantization techniques are extensively used to shrink LLMs' memory consumption. \citet{xiao2023smoothquant} apply rescaling between activations and parameters to eliminate outliers and simplify the quantization process. \citet{dettmers2022llm} breaks down matrix multiplications into predominantly 8-bit and a minority of 16-bit operations. \citet{frantar2022gptq} iteratively round weight columns into 3/4 bits, while \citet{lin2023awq} present an activation-aware quantization scheme to protect salient weights and compress LLMs to 3/4 bits. \citet{kim2023squeezellm} introduce a sparse plus low-precision pattern to handle a minor portion of vital weights, among other techniques.

\paragraph{Speculative Decoding.} As an approach orthogonal to the aforementioned methods, speculative decoding~\citep{leviathan2022fast,chen2023accelerating} aims to execute several decoding steps in parallel, thus reducing the total number of steps required. This parallelization is realized by employing a smaller draft model to conjecture several subsequent words, which the LLMs then collectively evaluate and accept as appropriate. While resonating with non-autoregressive generation literature~\citep{xiao2023survey}, this method is specifically tailored for LLMs to address the aforementioned inefficiency. Unlike previous works, we propose leveraging the original model to make predictions rather than introducing an additional draft model. This approach is more straightforward and seamlessly integrates into existing systems without the complexities of managing two models. Independently, \citet{miao2023specinfer, spector2023accelerating} propose the use of tree-structured attention to generate multiple candidates in parallel, where \citet{miao2023specinfer} suggest employing an ensemble of models to propose candidates, and \citet{spector2023accelerating} advocate adding another hierarchy for the draft model. 
\textcolor{black}{However, draft models require specialized pretraining and alignment with the target models. While employing multiple draft models can be cumbersome and involves the complexity of managing parallelism, our approach, which relies solely on decoding heads, offers a simpler alternative. \citet{miao2023specinfer} employ multiple draft models to generate tokens and merge them using tree attention, while \citet{spector2023accelerating} utilize a small draft model to process each level of the tree in batches. In contrast, our method directly uses the top predicted tokens from each of \ours heads to create a static sparse tree without autoregression or adjusting the tree structure. This approach simplifies the process and improves efficiency. Additionally, we demonstrate through a detailed ablation study how the nodes of the tree can affect decoding speed.}

\subsection{Sampling Scheme}
The manner in which text is sampled from Large Language Models (LLMs) can significantly influence the quality of the generated output. Recent studies have revealed that direct sampling from a language model may lead to incoherent or nonsensical results~\citep{pillutla2021mauve,holtzman2020curious}. In response to this challenge, \emph{truncation sampling} schemes have been introduced~\citep{fan2018hierarchical,basu2021mirostat,meister2022probability,hewitt2022truncation,meister2023locally}. These approaches aim to produce high-quality and diverse samples by performing sampling on a truncated distribution over a specific \emph{allowed set} at each decoding step.

Different strategies define this allowed set in various ways. For example, top-$k$ sampling~\citep{fan2018hierarchical} retains the $k$ most likely words, whereas top-$p$ sampling~\citep{holtzman2020curious} incorporates the minimal set of words that account for $p$ percent of the probability. Another method, known as typical decoding~\citep{meister2023locally}, employs the entropy of the predicted distribution to establish the threshold for inclusion. \citet{hewitt2022truncation} offers a unified framework to understand truncation sampling techniques comprehensively.

Drawing inspiration from these methods, our typical acceptance scheme aligns with the concept of defining an allowed set to exclude improbable candidates from the sampling process. However, we diverge because we do not insist on an exact correspondence between the output and language model distribution. This deviation allows us to facilitate more diverse yet high-quality outputs, achieving greater efficiency without compromising the integrity of the generated text.

\section{Experiment Settings}\label{appendix:experiment_settings}
\subsection{Common Terms}
We clarify three commonly used terms:
a) Acceleration rate: This refers to the average number of tokens decoded per decoding step. In a standard auto-regressive model, this rate is 1.0.
b) Overhead: This is used to characterize the per decoding step overhead compared to classic decoding, and is calculated by dividing the average per step latency of the \ours models by that of the vanilla model.
c) Speedup: This refers to the wall-time acceleration rate. 
Following these definitions, we have the relation: Speedup = Acceleration rate / Overhead.
\subsection{Shared Settings} 
For all the experiments, we use the Axolotl~\citep{axolotl2023} framework for training. We use a cosine learning rate scheduler with warmup and use 8-bit AdamW~\citep{dettmers20218bit} optimizer. We train $5$ \ours heads with $1$ layer and set $\lambda_k$ in Eq.~\eqref{eq:loss_medusa_1} to be $0.8^k$. For \ours-2, we use either LoRA~\citep{hu2021lora} or QLoRA~\citep{dettmers2023qlora} for fine-tuning and set the learning rate of \ours heads to be $4$ times larger than the backbone model. LoRA is applied to all the linear layers of the backbone model, including the language model head. The rank of LoRA adapter is set to $32$, and $\alpha$ is set to $16$. A dropout of $0.05$ is added to the LoRA adapter. 

\subsection{\ours-1 v.s. \ours-2 on Vicuna 7B and 13B} 
We use a global batch size of $64$ and a peak learning rate of $5e^{-4}$ for the backbone and $2e^{-3}$ for \ours heads and warmup for $40$ steps. We use $4$-bit quantized backbone models for both models. We first train the models with \ours-1 and use these trained models as initialization to train \ours-2. We employ QLoRA for \ours-2 and the  $\lambda_0$ in Eq.~\eqref{eq:loss_medusa_2} is set to be $0.2$.
\subsection{ Training with Self-Distillation on Vicuna-33B and Zephyr-7B} 
We use \ours-2 for both models instead of using a two-stage training procedure. We use a sine schedule for the $\theta_0$ to gradually increase the value to its peak at the end of the training. We find this approach is equally effective. We set the peak learning rate of the backbone LoRA adapter to be $1e^{-4}$ and the warmup steps to be $20$ since the self-distillation loss is relatively small. We set the $\lambda_0$ in Eq.~\eqref{eq:loss_medusa_2} to be $0.01$.
\section{Visualization of optimized tree attention}\label{appendix:sparse_tree}
\begin{figure*}[h]
    \centering
    \includegraphics[width=0.6\textwidth]{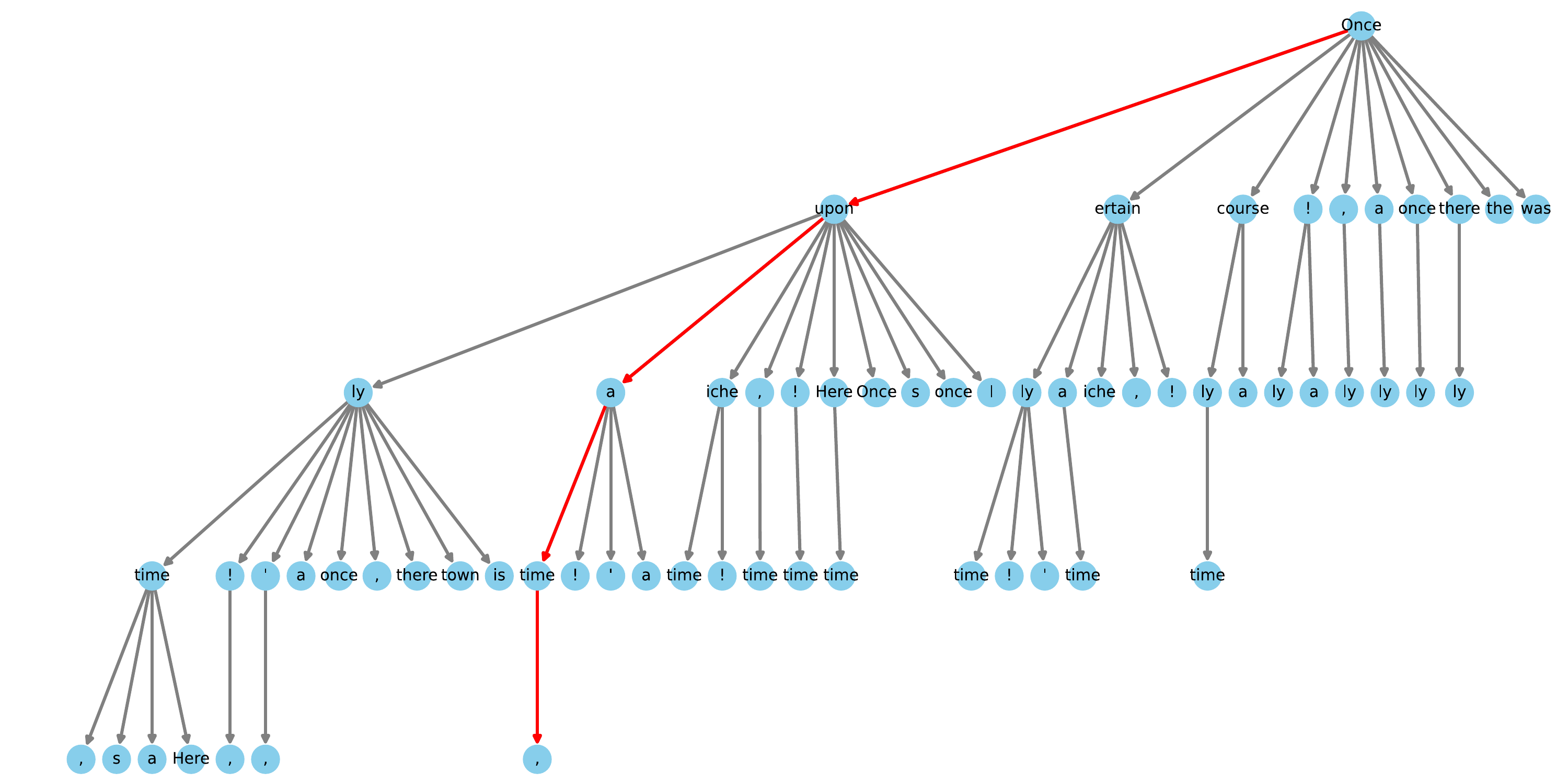}
    \caption{Visualization of a sparse tree setting for \ours-2 Vicuna-7B. The tree has \textcolor{black}{64 nodes representing candidate tokens} and a depth of 4 which indicates 4 \ours heads involved in calculation. Each node indicates a token from a top-k prediction of a \ours head, and the edges show the connections between them. The red lines highlight the path that correctly predicts the future tokens.}
    \label{fig:sparse_tree}
\end{figure*}

Fig.~\ref{fig:sparse_tree} illustrates the structure of a sparsely constructed tree for the \ours-2 Vicuna-7B model. This tree structure extends four levels deep, indicating the engagement of four \ours heads in the computation. The tree is initially formed through a Cartesian product approach and subsequently refined by pruning based on the statistical expectations of the top-k predictions from each \ours head measured on the Alpaca-eval dataset~\cite{dubois2023alpacafarm}. The tree's lean towards the left visually represents the algorithm's preference for nodes with higher probabilities on each head.

\section{Results of Speculative Decoding}\label{appendix:spec}

In this study, speculative decoding was applied to Vicuna models~\citep{vicuna2023} with varying sizes, specifically 7B, 13B, and 33B. The preliminary framework utilized open-source models such as Llama-68M and 160M~\citep{miao2023specinfer}, alongside Tiny-Llama~\citep{zhang2024tinyllama} and Tiny-Vicuna~\citep{tiny_vicuna_1b}, fine-tuned from Tiny-Llama with the Vicuna-style instructional tuning strategy. Due to the proprietary nature of speculative decoding methods~\citep{chen2023accelerating, leviathan2022fast}, open-source alternatives\footnote{\href{https://github.com/feifeibear/LLMSpeculativeSampling}{https://github.com/feifeibear/LLMSpeculativeSampling}} were deployed for evaluation. Additionally, we utilize \verb|torch.compile()| to accelerate the inference speed of draft models.

\begin{figure*}[h]
     \centering
     \begin{subfigure}[b]{0.32\textwidth}
         \centering
         \includegraphics[width=\textwidth]{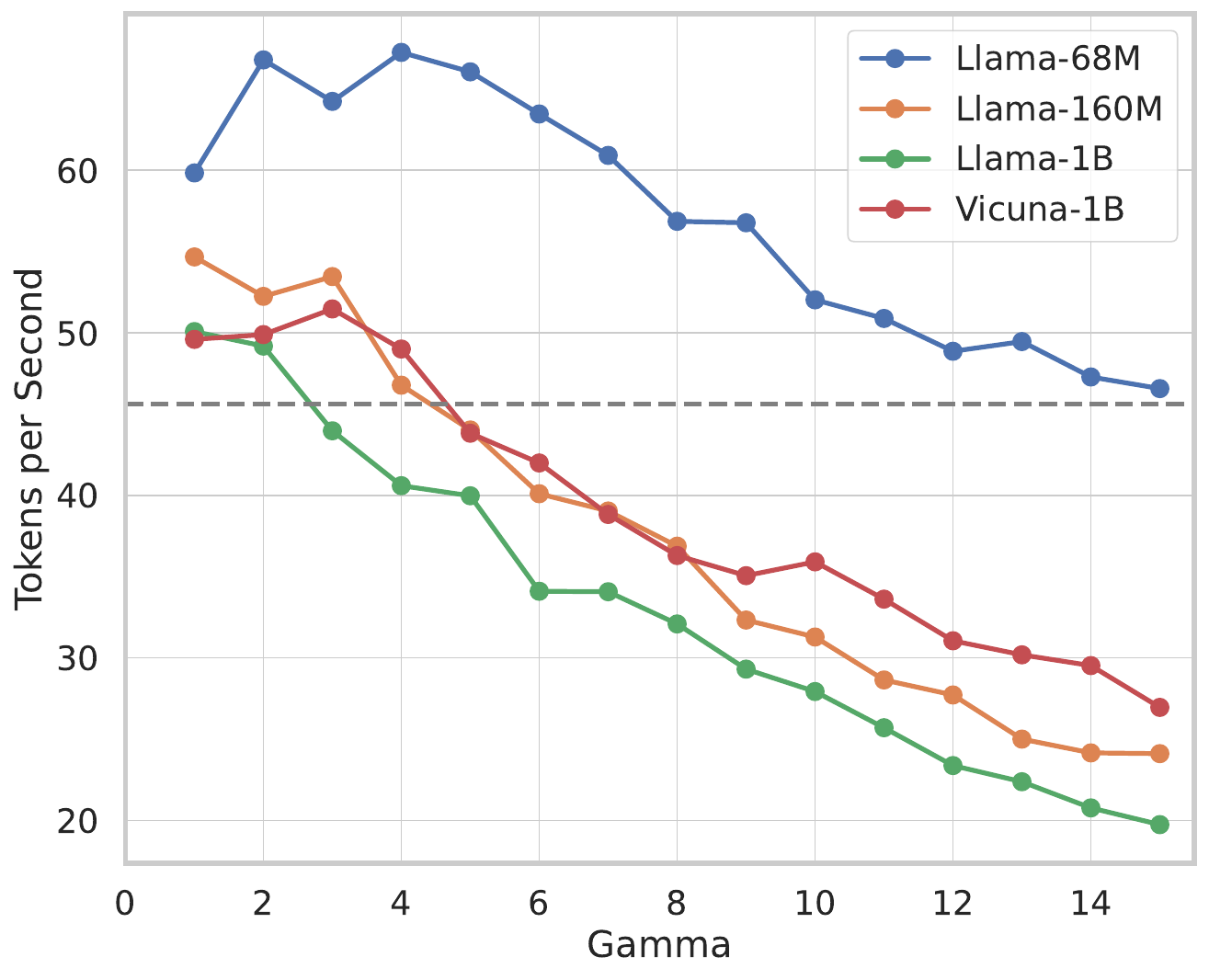}
         \caption{Vicuna-7B}
         \label{fig:spec7b}
     \end{subfigure}
     \begin{subfigure}[b]{0.32\textwidth}
         \centering
         \includegraphics[width=\textwidth]{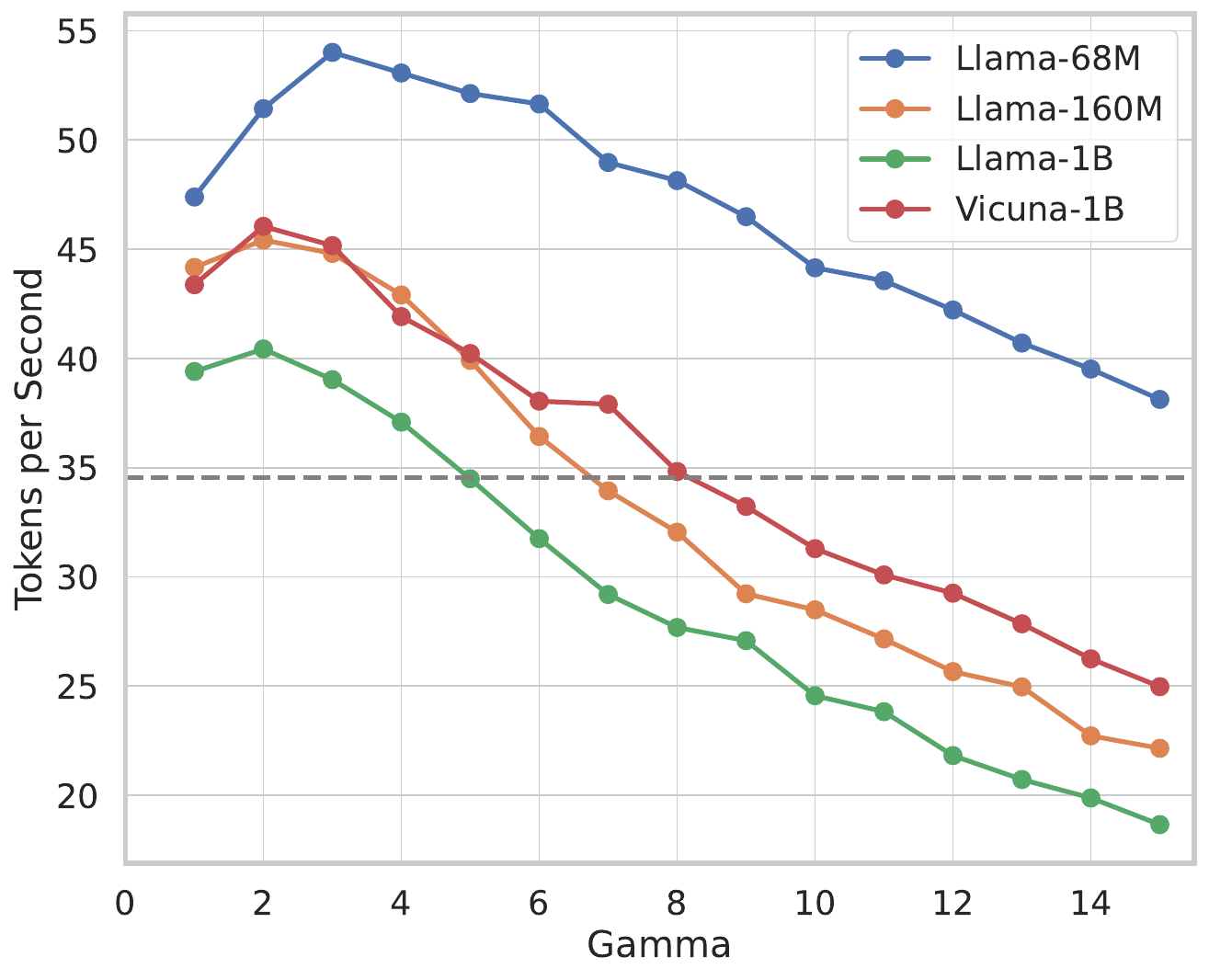}
         \caption{Vicuna-13B}
         \label{fig:spec13b}
     \end{subfigure}
    \begin{subfigure}[b]{0.32\textwidth}
         \centering
         \includegraphics[width=\textwidth]{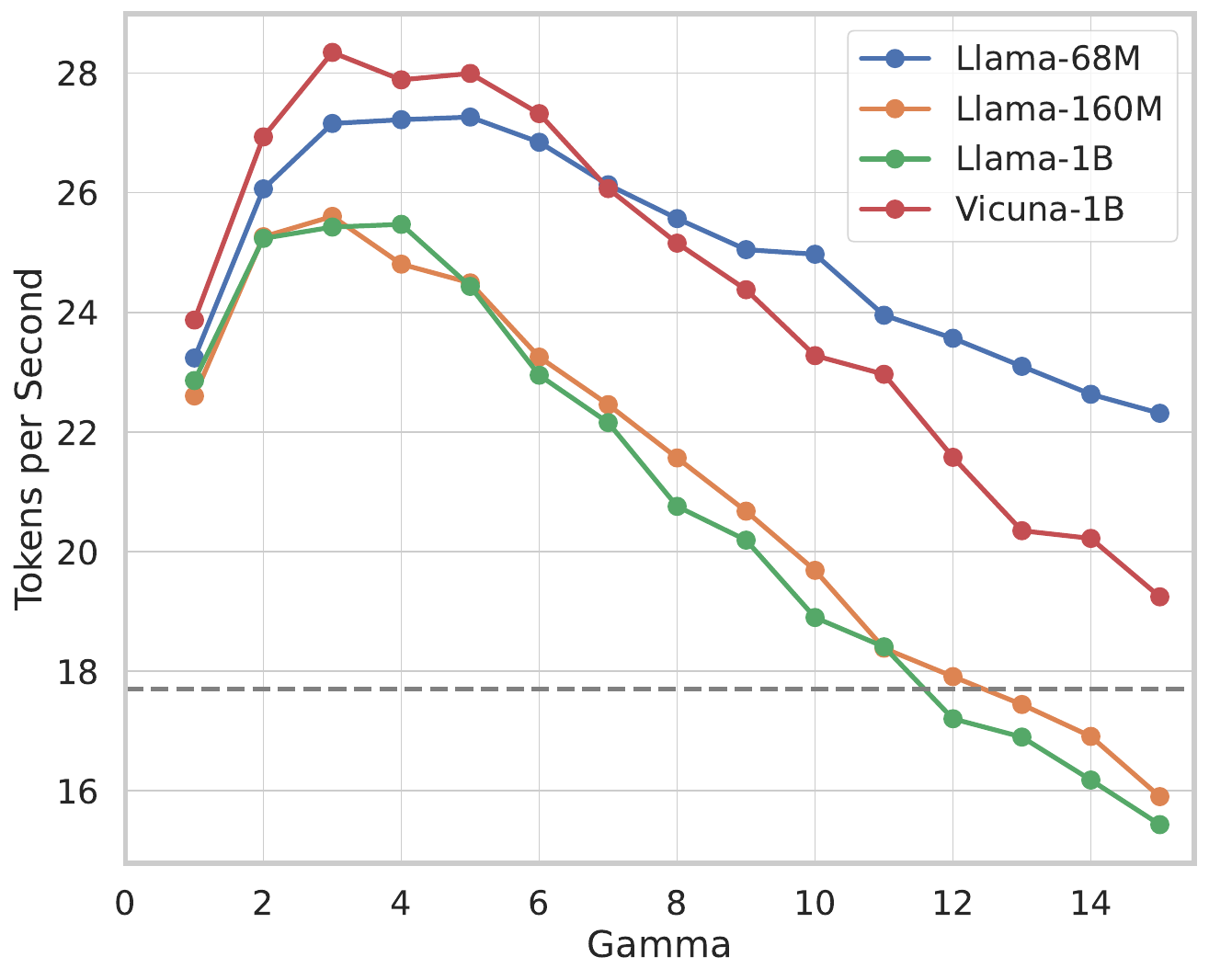}
         \caption{Vicuna-33B}
         \label{fig:spec33b}
     \end{subfigure}
        \caption{Inference speed of various models using speculative decoding on MT-Bench. Baseline model speeds are presented by grey dotted lines for comparison. $\gamma$ denotes the draft token number.}
        \label{fig:speculative_decoding}
\end{figure*}

Our results shown in Fig.~\ref{fig:speculative_decoding}, reveal that the optimal settings of the draft model vary with the Vicuna model sizes. Specifically, the Llama-68M, with a setting of the draft token number $\gamma=4$, yielded the best performance for Vicuna-7B, while the same draft model with $\gamma=3$ was most effective for Vicuna-13B. For the larger Vicuna-33B, the Tiny-Vicuna \textcolor{black}{(Vicuna-1B)}, with $\gamma=3$, provided the greatest acceleration. These results suggest that the choice and setting of the drafting model should be tailored to the size of the LLMs, presenting an area for further exploration in the field.

\section{Additional Results for All Models}\label{appendix:add_results}
\begin{figure}[h]
    \centering
    \includegraphics[width=0.45\textwidth]{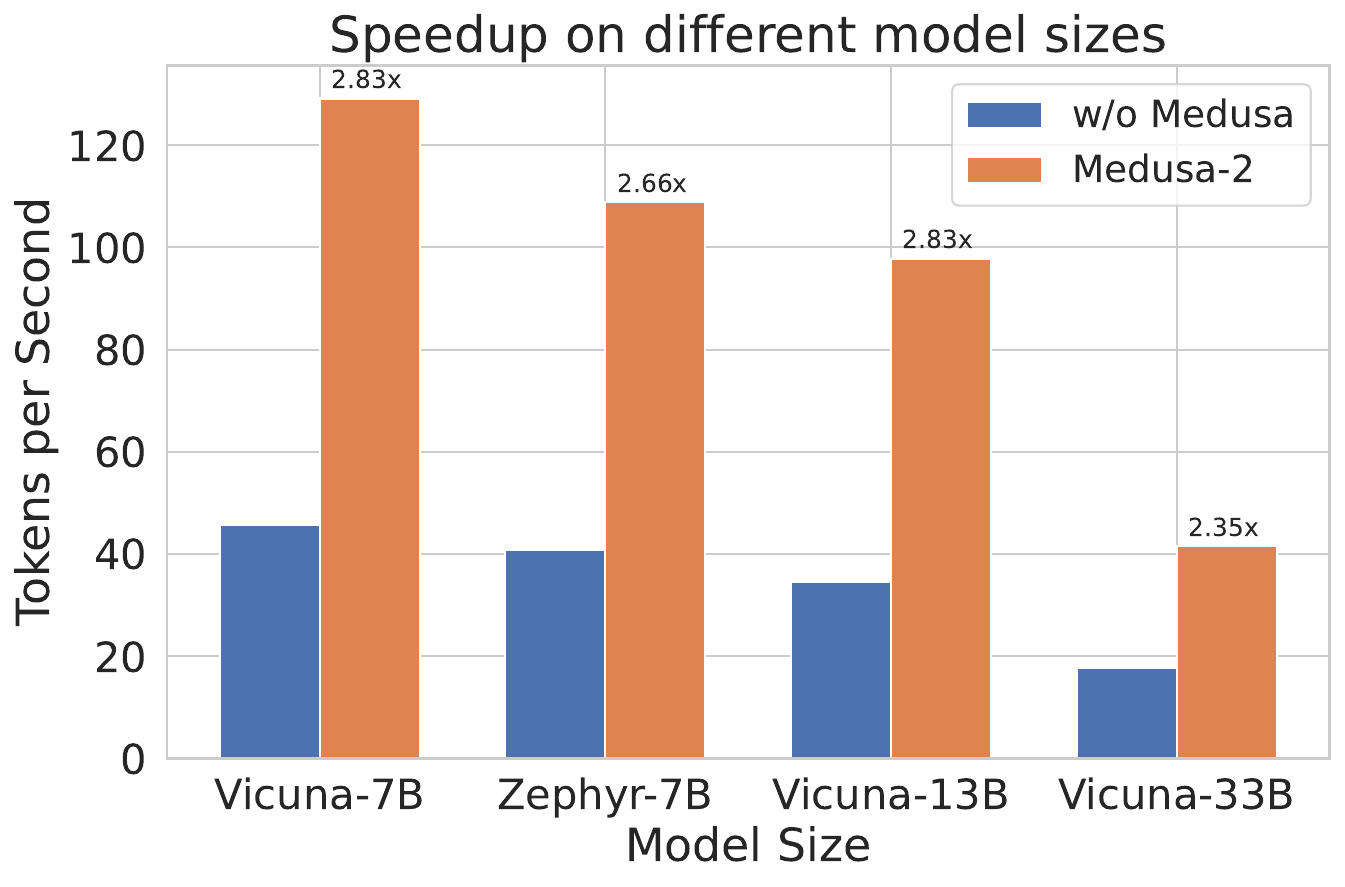}
    \caption{Speedup of various models with \ours-2. \ours-2 shows significant speed improvement over all the models, while models trained with self-distillation \textcolor{black}{(Zephyr-7B, Vicuna-13/33B)} have weaker speedup due to the trade-off between preserving quality and boosting speed.}
    \label{fig:speedup_model_wild}
\end{figure}

We show speedup on various models in Fig.~\ref{fig:speedup_model_wild}.

\section{Additional Results on AlpacalEval Dataset}
We conduct further experiments on the AlpacaEval~\citep{alpaca_eval} dataset. \ours-2 achieves consistent speedup similar to the results on MT-Bench.
\begin{table}[h]
    \centering
    \begin{tabular}{llrrrr}
    \toprule
     & Model & Base speed (tokens/s) & \ours speed (tokens/s) & Acc. rate & Speedup \\
    \midrule
     & Vicuna-7b & 37.07 & 106.76 & 3.23 & 2.88 \\
     & Vicuna-13b & 29.01 & 91.54 & 3.28 & 3.16 \\
     & Vicuna-33b & 17.87 & 40.43 & 2.85 & 2.26 \\
     & Zephyr-7b & 34.21 & 99.50 & 3.08 & 2.91 \\
    \bottomrule
    \end{tabular}
    \caption{Speedup results on AlpacaEval~\citep{alpaca_eval} dataset.}
    \label{tab:alpaca_eval_speedup}
\end{table}

\section{Exploration and Modeling of Hardware Constraints and \ours}~\label{sec:roofline}

We explore the hardware constraints, specifically memory-bandwidth bound, and their impact on \ours-style parallel decoding by incorporating a simplified \textcolor{black}{Llama-series} model.
First, we \textcolor{black}{identify} that the operators involving matrix multiplications, such as linear layers and attention matrix multiplications, are the primary sources of overhead. We profile the performance of FLOP/s vs. Operational Intensity \textcolor{black}{which is the ratio of FLOP/s to bandwidth (bytes/s)}, across various GPUs, including the A100-80GB-PCIe, A40, and A6000.
Next, we examine the changes in FLOP/s vs. Operational Intensity when using \ours for different operators.
Finally, we apply a straightforward analytical model to calculate acceleration rates and combine it with hardware benchmarks. This provides insights into the effects under different model sizes, sequence lengths, and batch sizes.

\subsection{Roofline Model of Operators}
We present an analysis of the roofline model for various operators in large language models (LLMs), specifically focusing on Llama-7B, Llama-13B, and Llama-33B~\cite{touvron2023llama}. These models were benchmarked on different GPUs, including the A100-80GB-PCIe, A40, and A6000. We looked into the three categories of matrix multiplication operators since they represent the primary sources of computational overhead in these models. Our study follows the report~\cite{chen2023transformer} which investigates the effectiveness of batch size but ours focuses more on decoding and parallel decoding.

Table~\ref{tab:complexity} details the computation and space complexity for each operator during the prefill, decoding, and \ours decoding phases. The operators include the linear layers for query, key, and value matrices ($XW_{Q}$, $XW_{K}$, $XW_{V}$), the attention matrix multiplications ($QK^T$, $PV$), and the up/gate/down linear layers ($XW_{u}$, $XW_{g}$, $XW_{d}$).
$b$ stands for the batch size, $s$ stands for the sequence length, $h$ stands for the hidden dimension, $i$ stands for the intermediate dimension, $n$ stands for the number of attention heads, $d$ stands for the head dimension and $q$ stands for the candidate length for \ours.
For more details of these operators please refer to the articles~\cite{touvron2023llama, chen2023transformer}.

\begin{table}[h]
\centering
\caption{Computational and space complexity of the main operators in different phases. \textcolor{black}{The table is based on Table 2 in the report~\cite{chen2023transformer}.}}

\scriptsize
\begin{tabular}{lcccc}
\toprule
 \textbf{Operator} & \textbf{Input Shape} & \textbf{Output Shape} & \textbf{Comp. Complexity} & \textbf{Space Complexity} \\ \midrule
 \textbf{Prefill} \\ \midrule
 $XW_{Q}$, $XW_{K}$, $XW_{V}$ & $(b, s, h)$ & $(b, s, h)$ & $O(bsh^2)$ & $O(2bsh + h^2)$ \\ \midrule
  $QK^T$ & $(b, n, s, d),(b, n, s, d)$ & $(b, n, s, s)$ & $O(bs^2nd)$ & $O(2bsnd + bs^2n)$ \\ 
  $PV$ &$(b, n, s, s),(b, n, s, d)$&$(b, n, s, d)$&& \\ \midrule
  $XW_{u}$, $XW_{g}$ & $(b, s, h)$ & $(b, s, i)$ & $O(bshi)$ & $O(bs(h + i) + hi)$ \\ 
  $XW_{d}$&$(b, s, i)$&$(b, s, h)$&&\\ \midrule
   \textbf{Decoding} \\ \midrule
$XW_{Q}$, $XW_{K}$, $XW_{V}$ & $(b, 1, h)$ & $(b, 1, h)$ & $O(bh^2)$ & $O(2bh + h^2)$ \\ \midrule
  $QK^T$ & $(b, n, 1, d), (b, n, s, d)$ & $(b, n, s, 1)$ & $O(bsnd)$ & $O(bsn + bsnd + bnd)$ \\ 
  $PV$ & $(b, n, s, 1), (b, n, 1, d)$ & $(b, n, 1, d)$ & &  \\ \midrule
  $XW_{u}$, $XW_{g}$ & $(b, 1, h)$ & $(b, 1, i)$ & $O(bhi)$ & $O(b(h + i) + hi)$ \\
  $XW_{d}$ & $(b, 1, i)$ & $(b, 1, h)$ &  & \\\midrule
   \textbf{Parallel decoding} \\ \midrule
 $XW_{Q}$, $XW_{K}$, $XW_{V}$ & $(b, q, h)$ & $(b, q, h)$ & $O(bqh^2)$ & $O(2bqh + h^2)$ \\ \midrule
  $QK^T$ & $(b, n, q, d), (b, n, s, d)$ & $(b, n, s, q)$ & $O(bsqnd)$ & $O(bsqn + b(s+q)nd)$ \\ 
    $PV$ & $(b, n, s, q), (b, n, q, d)$ & $(b, n, q, d)$ & &  \\ \midrule
  $XW_{u}$, $XW_{g}$ & $(b, q, h)$ & $(b, q, i)$ & $O(bqhi)$ & $O(bq(h + i) + hi)$ \\
  $XW_{d}$ & $(b, q, i)$ & $(b, q, h)$ &  \\ \bottomrule
\end{tabular}
\label{tab:complexity}
\end{table}

\begin{figure}[h]
    \centering
    \includegraphics[width=0.8\textwidth]{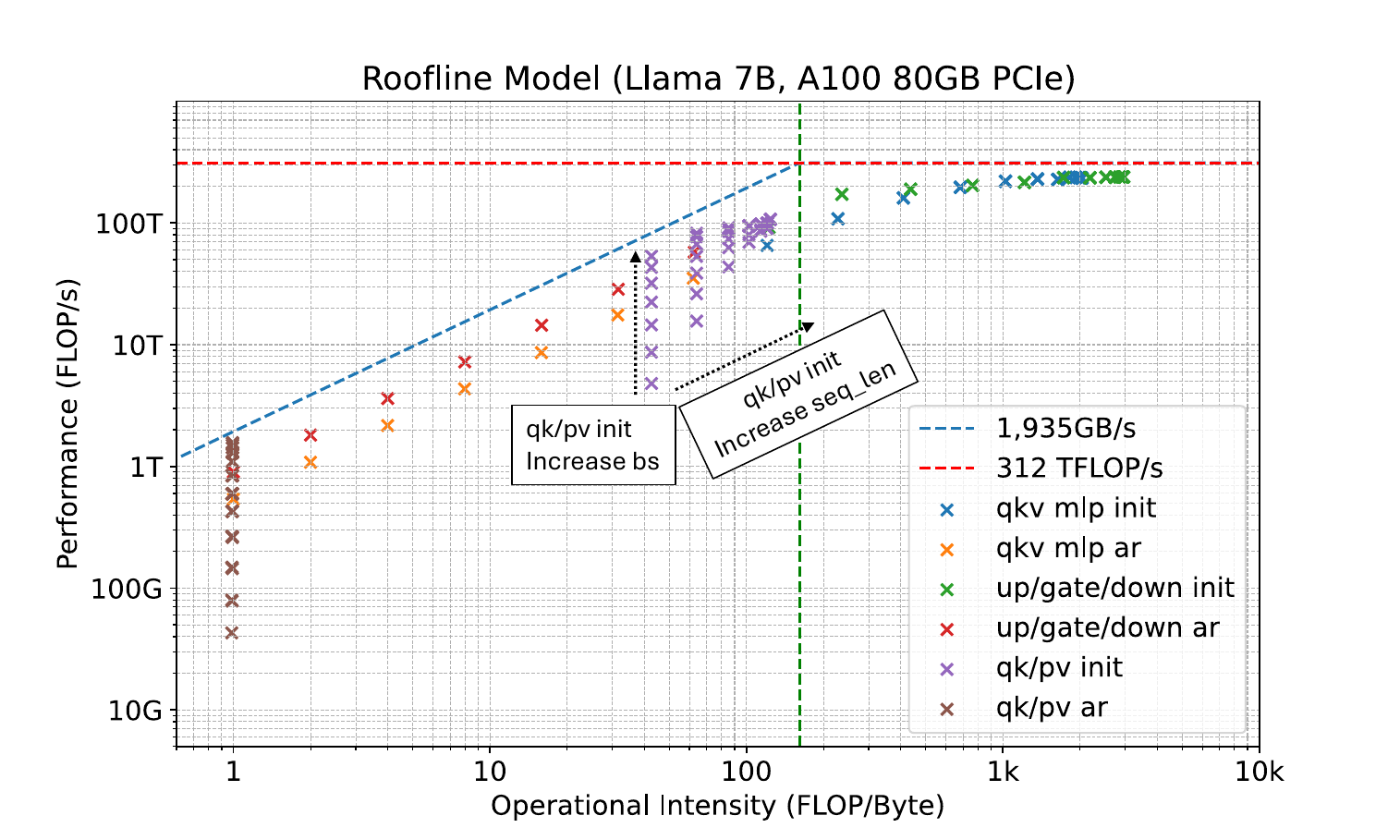}
    \caption{The figure shows the relationship between FLOP/s and Operational Intensity for all benchmarked datapoints of Llama-7B operators on A100-80GB-PCIe. The dashed lines represent the HBM bandwidth limit (1,935GB/s) and the peak performance limit (312 TFLOP/s)~\cite{nvidia_a100_datasheet}. `\texttt{qkv mlp}' stands for the linear layers projecting hidden features to query/key/value features. `\texttt{up/gate/down}' stands for the linear layers following the attention block. `\texttt{qk/pv}' stands for the two steps of attention matrix multiplications. `\texttt{ar}' stands for the decoding (autoregressive) and `\texttt{init}' stands for the prefill phase.}
    \label{fig:llama7b-roofline-a100}
\end{figure}

\begin{figure}[h]
    \centering
    \includegraphics[width=0.8\textwidth]{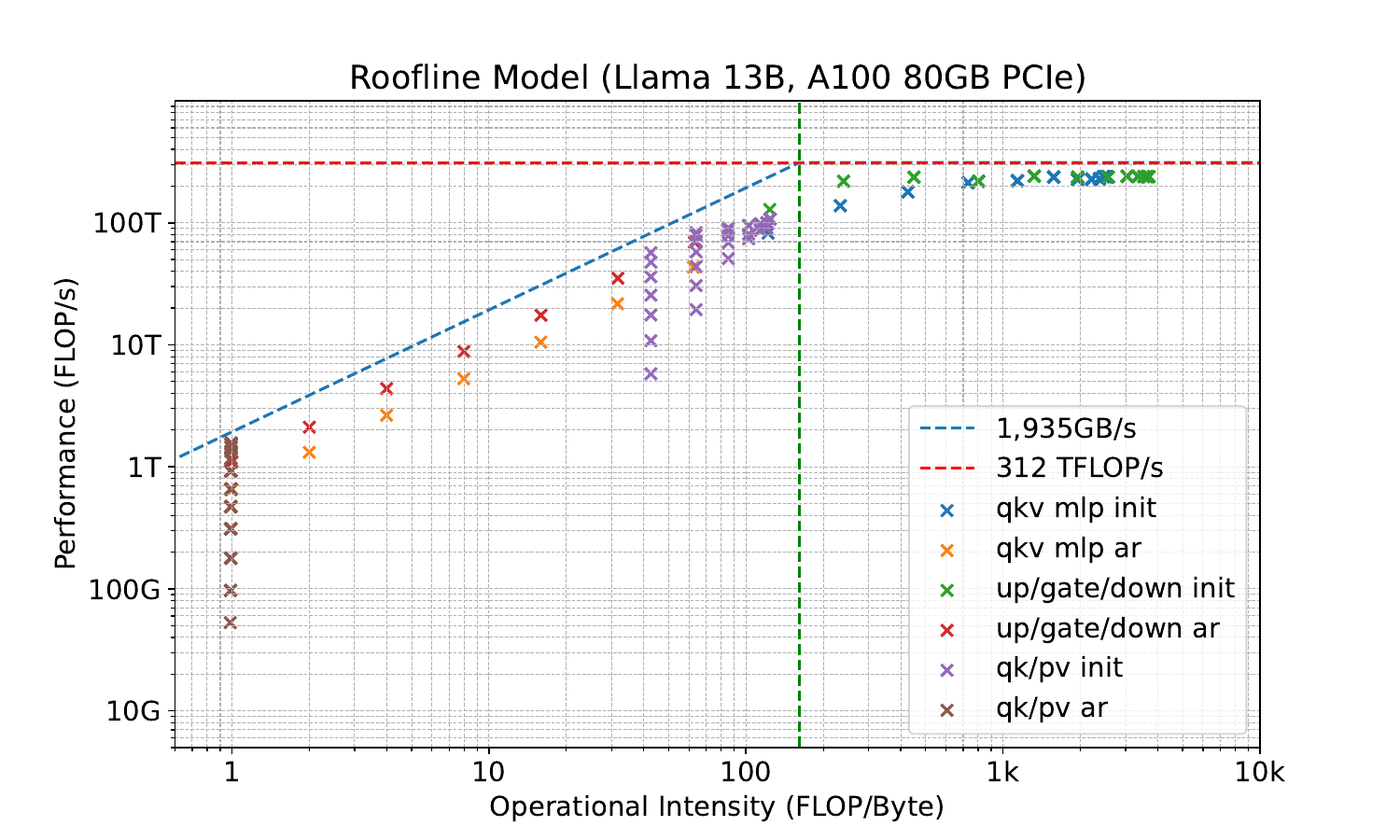}
    \caption{Llama-13B operators on A100-80GB-PCIe.}
    \label{fig:llama13b-roofline-a100}
\end{figure}

\begin{figure}[h]
    \centering
    \includegraphics[width=0.8\textwidth]{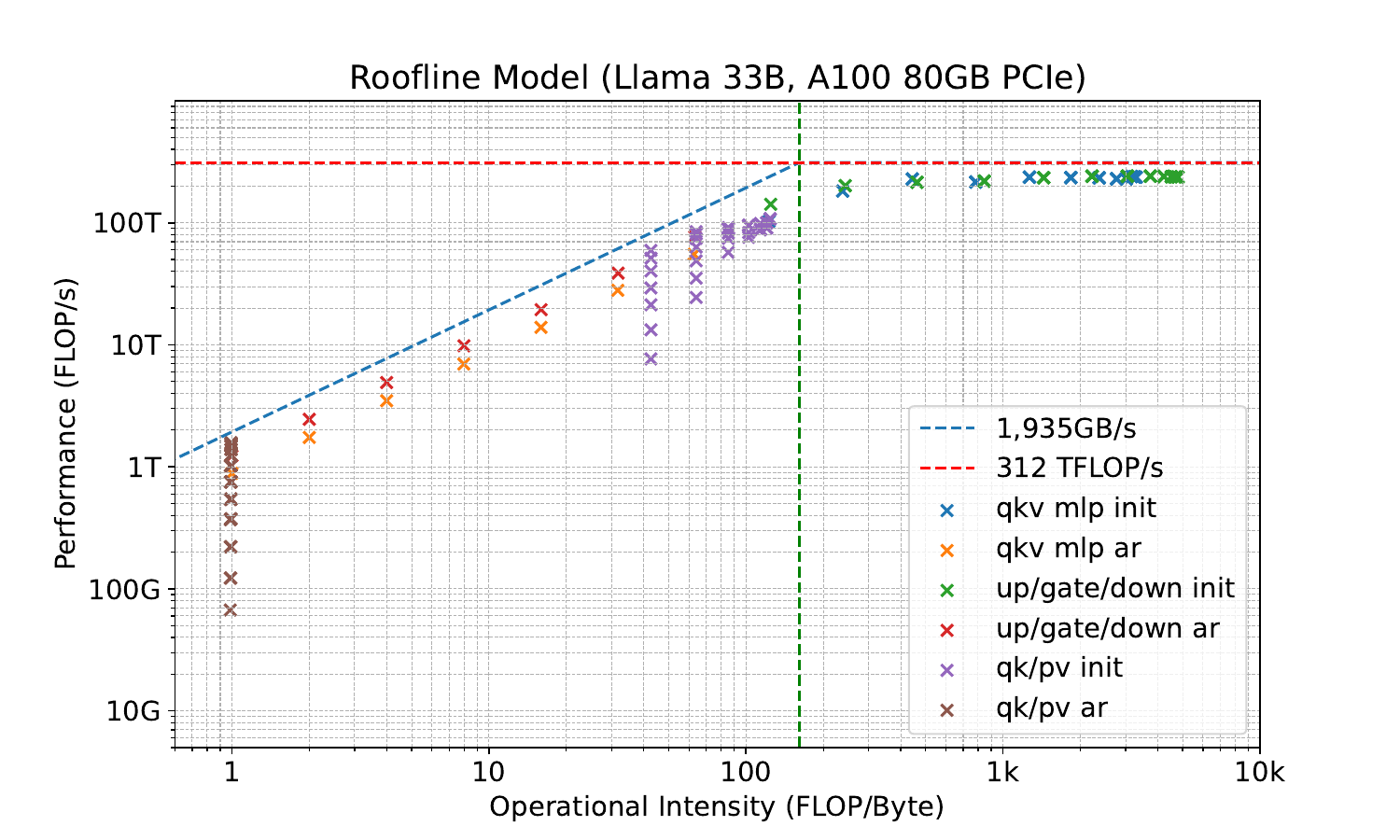}
    \caption{Llama-33B operators on A100-80GB-PCIe.}
    \label{fig:llama33b-roofline-a100}
\end{figure}

\begin{figure}[h]
    \centering
    \includegraphics[width=0.8\textwidth]{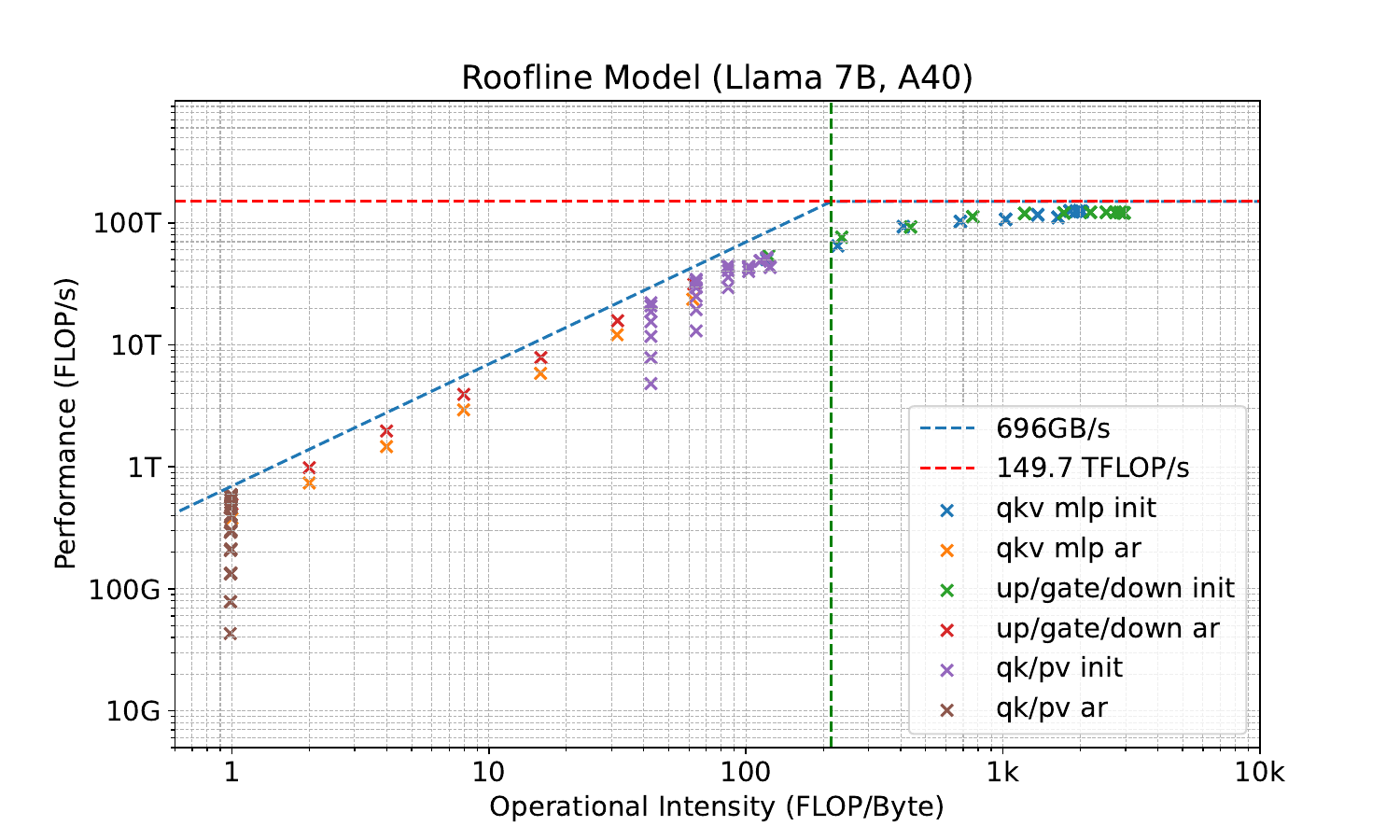}
    \caption{Llama-7B operators on A40.}
    \label{fig:llama7b-roofline-a40}
\end{figure}

\begin{figure}[h]
    \centering
    \includegraphics[width=0.8\textwidth]{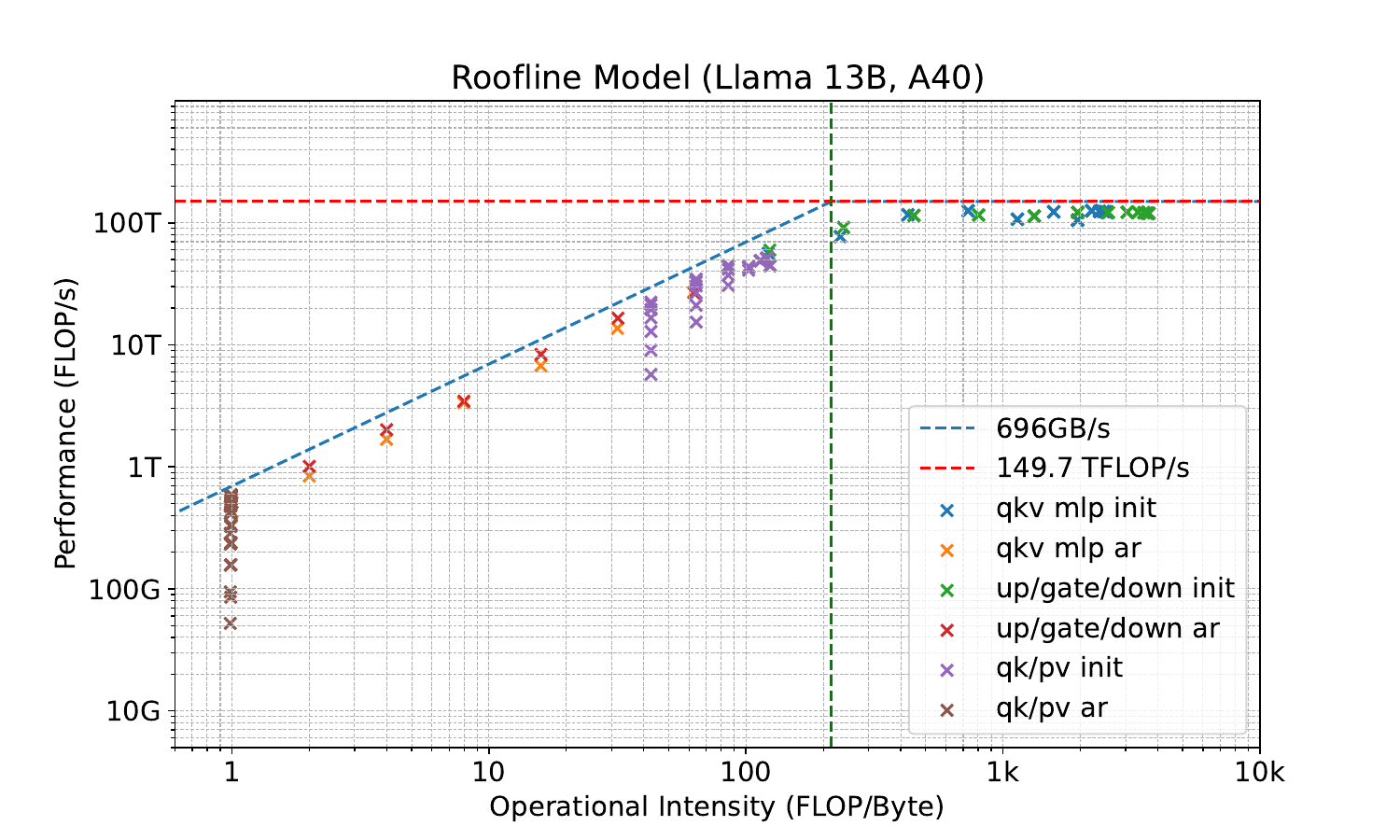}
    \caption{Llama-13B operators on A40.}
    \label{fig:llama13b-roofline-a40}
\end{figure}

\begin{figure}[h]
    \centering
    \includegraphics[width=0.8\textwidth]{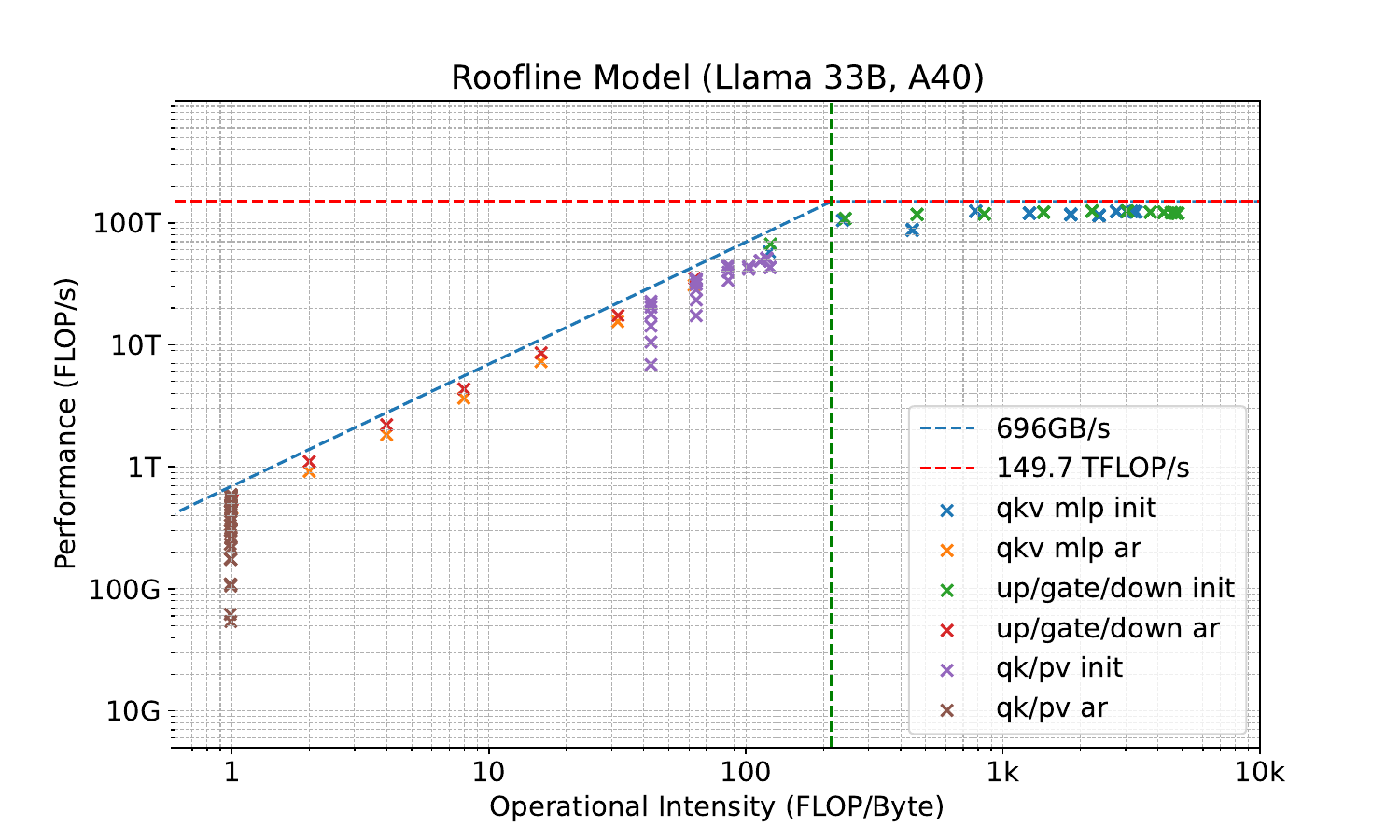}
    \caption{Llama-33B operators on A40.}
    \label{fig:llama33b-roofline-a40}
\end{figure}

\begin{figure}[h]
    \centering
    \includegraphics[width=0.8\textwidth]{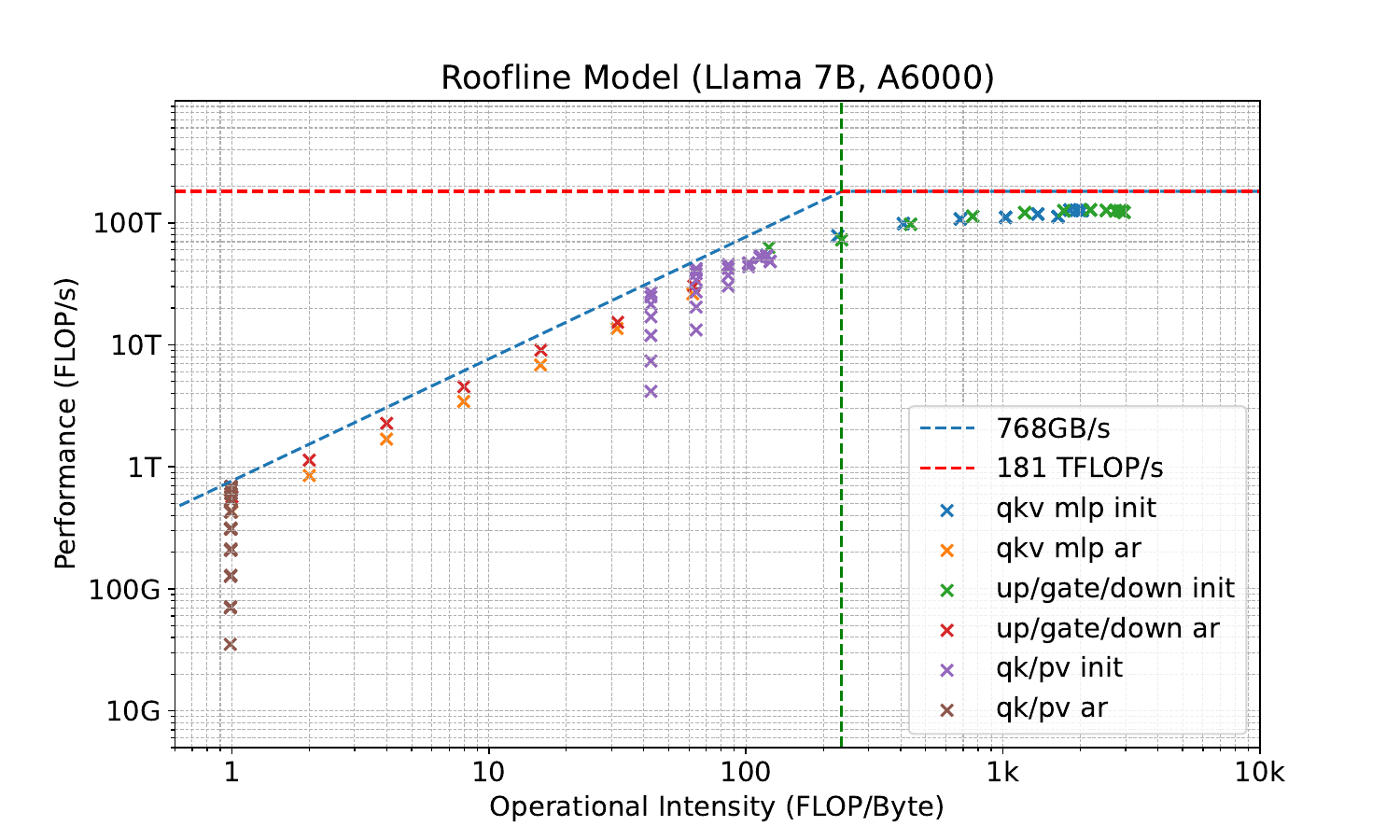}
    \caption{Llama-7B operators on A6000.}
    \label{fig:llama7b-roofline-a6000}
\end{figure}

\begin{figure}[h]
    \centering
    \includegraphics[width=0.8\textwidth]{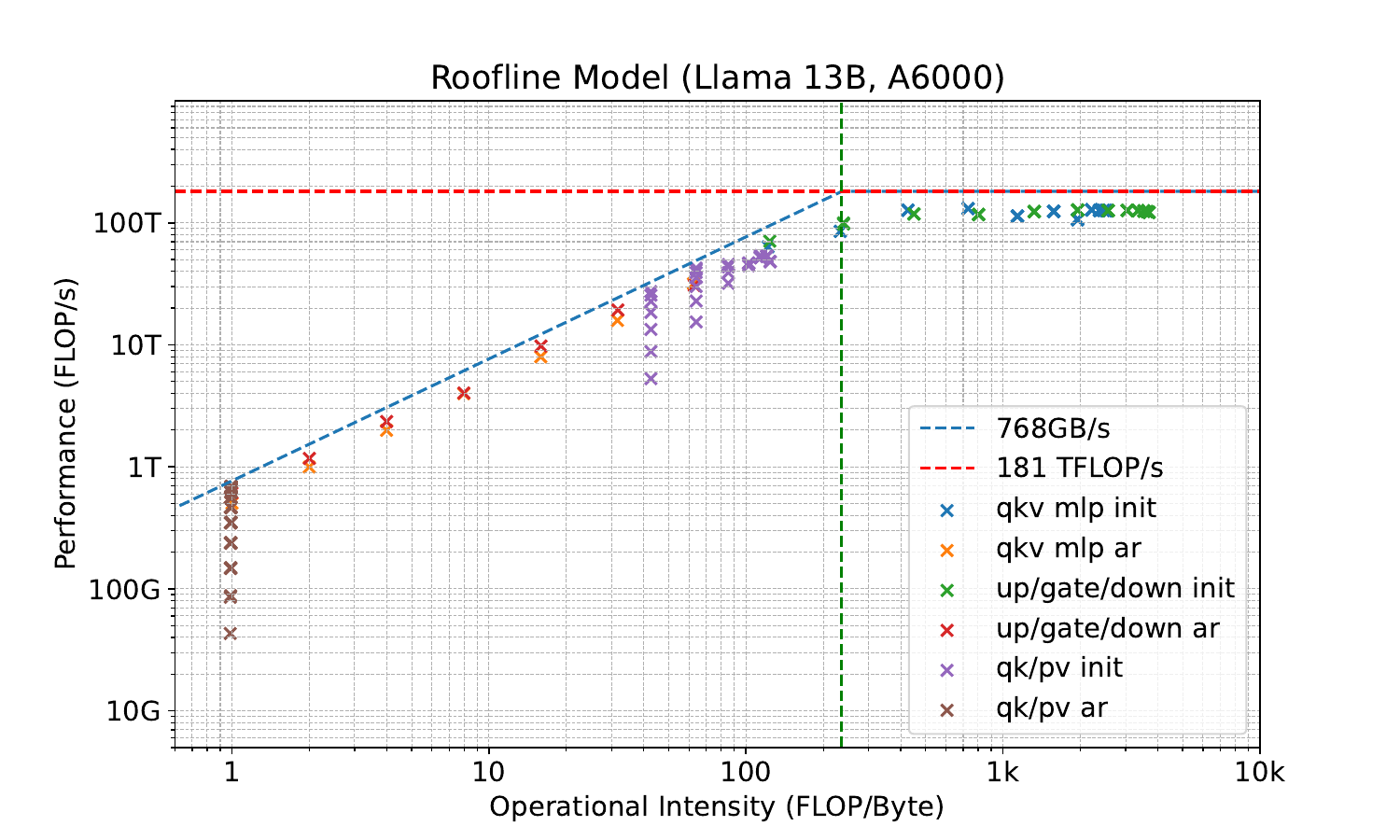}
    \caption{Llama-13B operators on A6000.}
    \label{fig:llama13b-roofline-a6000}
\end{figure}

\begin{figure}[h]
    \centering
    \includegraphics[width=0.8\textwidth]{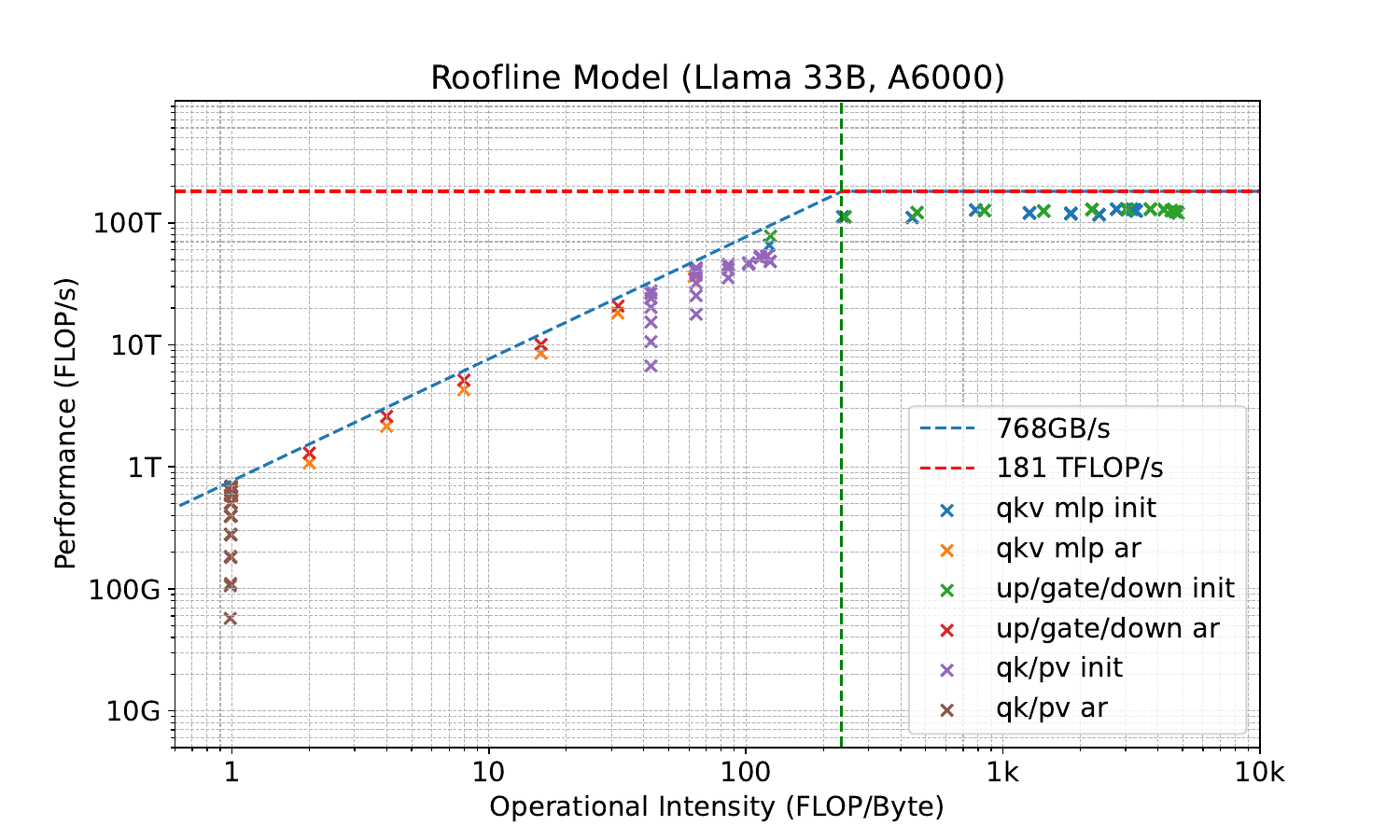}
    \caption{Llama-33B operators on A6000.}
    \label{fig:llama33b-roofline-a6000}
\end{figure}

Figures~\ref{fig:llama7b-roofline-a100}-\ref{fig:llama33b-roofline-a6000} show the benchmark of three categories of operators on different models (7/13/33B) under various settings. To evaluate each operator's performance and throughput, we chose the combination of settings including batch sizes from 1 to 64 in powers of 2 and sequence lengths from 128 to 8192 in powers of 2 \textcolor{black}{(49 settings for each operator)}. 
From all the figures, we observe that the datapoints of each operator in the prefill and decoding stages cluster at very similar positions across all GPUs and for various model sizes.

During the prefill phase, increasing the batch size changes the FLOP/s of the attention matrix multiplications (see \texttt{`qk/pv init`}) but does not affect the Operational Intensity (refer to the vertical dashed arrow in Fig. 9). 
In contrast, increasing the sequence length impacts both FLOP/s and Operational Intensity in the prefill phase (refer to the diagonal dashed arrow in Fig. 9).
During the decoding phase, the attention matrix multiplications are significantly limited by memory bandwidth. Despite an increase in FLOP/s with changes in batch size and sequence length, the Operational Intensity remains nearly unchanged (see \texttt{`qk/pv ar`}). This indicates suboptimal resource utilization in the self-attention mechanism.

The linear layers in the prefill phase are mostly compute-bound (see \texttt{`qkv mlp init`} and \texttt{`up/gate/down init`}). During the decoding phase, the datapoints of the linear layer form a line with the same slope as the GPU’s memory bandwidth (see \texttt{`qkv mlp ar`} and \texttt{`up/gate/down ar`}). This indicates the linear layers in the decoding stage are also bounded by memory bandwidth. Increasing the batch size improves the achieved FLOP/s and Operational Intensity under memory bandwidth constraints through better parallelism. Note that linear layers only process the new token and are independent of sequence length (See `Decoding` section in Table~\ref{tab:complexity}).

\clearpage

\subsection{FLOP/s vs. Operational Intensity Variations in \ours}

We investigate how Medusa can change Operational Intensity and elevate the FLOP/s.
We choose Llama 33B on A100-80GB-PCIe as the setting. 

\begin{figure}[h]
    \centering
    \includegraphics[width=0.8\textwidth]{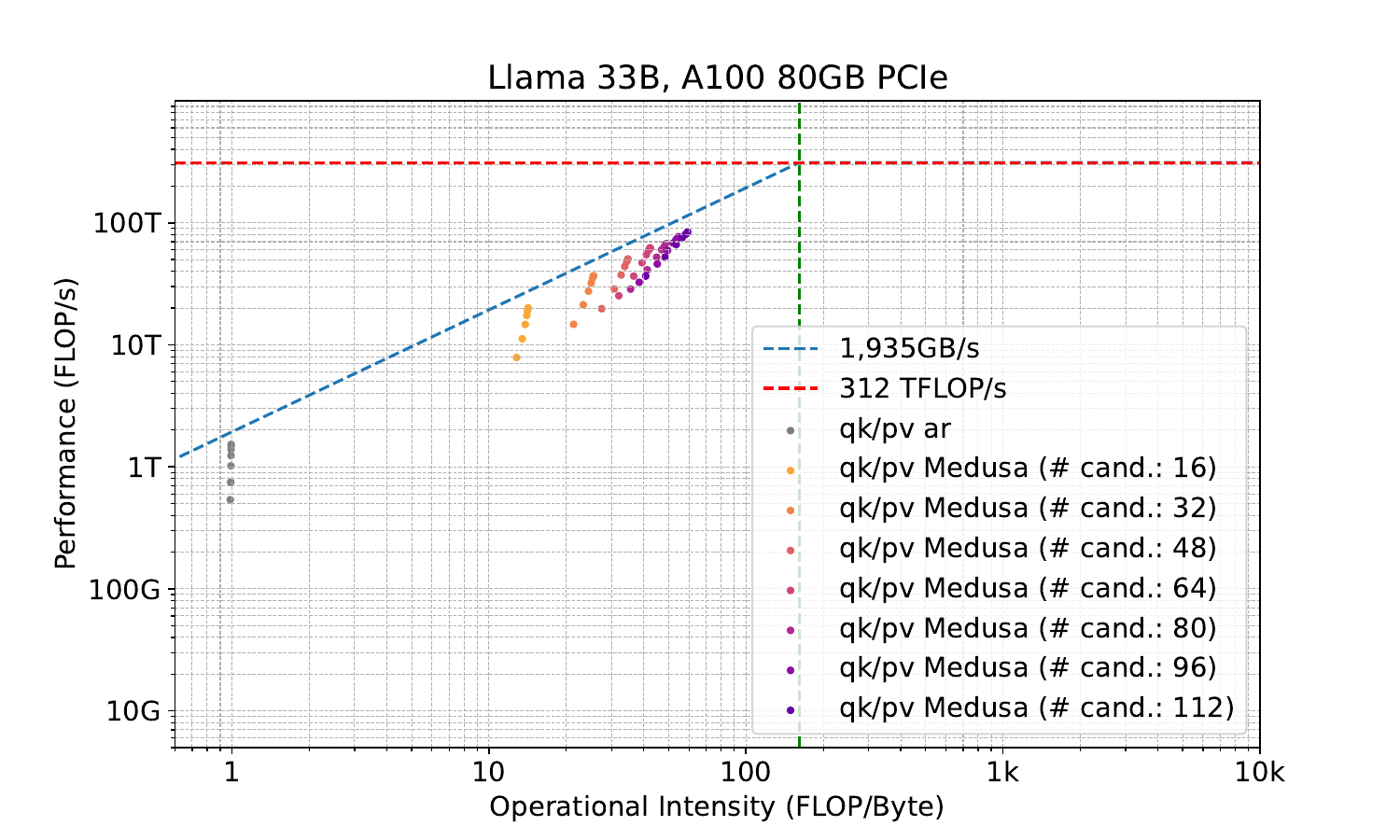}
    \caption{FLOP/s vs. Operational Intensity of attention matrix multiplication with batch size 16.}
    \label{fig:llama33b-spec-bs16}
\end{figure}

\begin{figure}[h]
    \centering
    \includegraphics[width=0.8\textwidth]{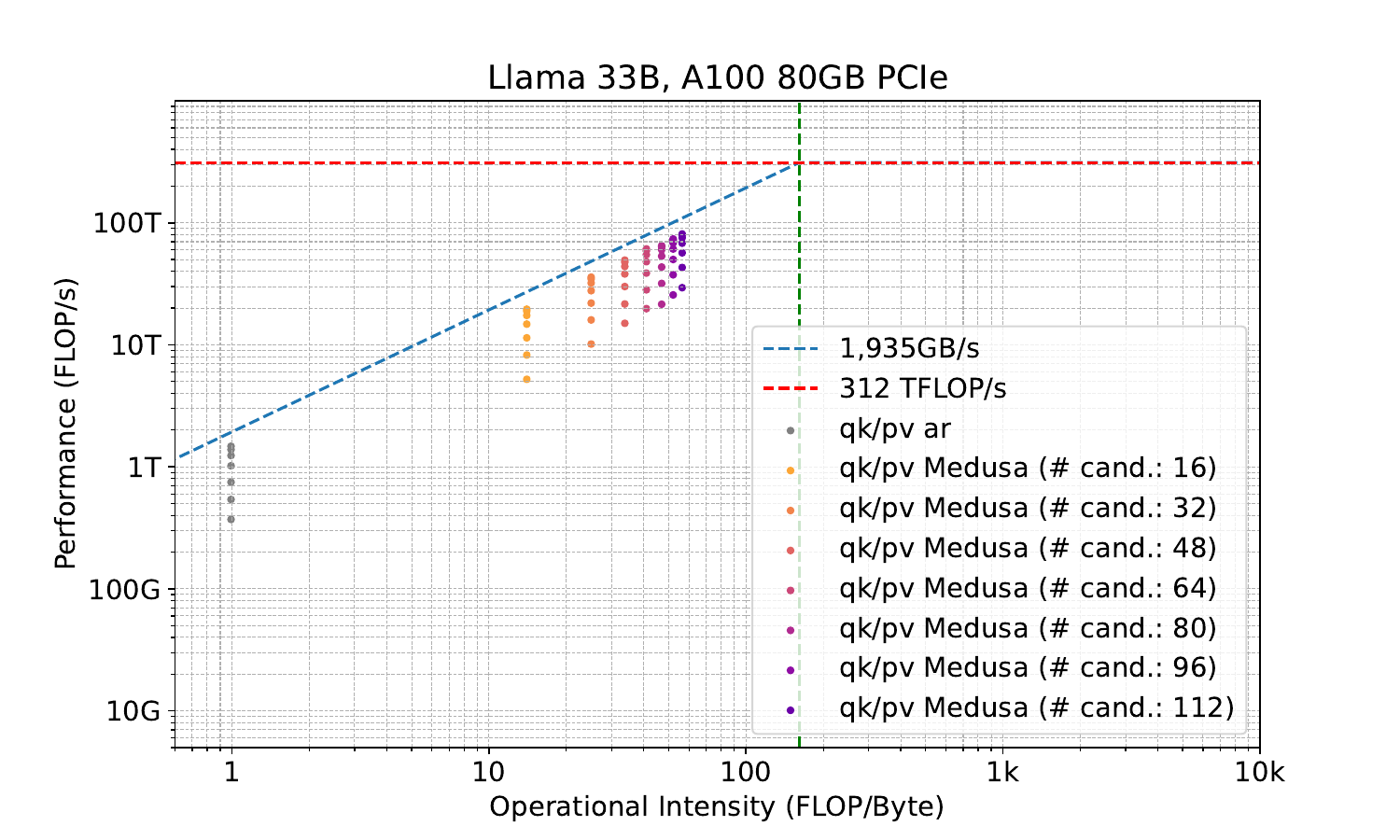}
    \caption{FLOP/s vs. Operational Intensity of attention matrix multiplication with sequence length 1024.}
    \label{fig:llama33b-spec-seq1024}
\end{figure}

First, we examine the attention matrix multiplication. Fig.~\ref{fig:llama33b-spec-bs16} and Table~\ref{tab:llama33b-spec-bs16} illustrate the effects of \ours while keeping the batch size fixed at 16. We observe increased FLOP/s and Operational Intensity as more candidate tokens are added (original decoding results are plotted as grey dots). This indicates that \ours can leverage additional candidate tokens to improve computational throughput. Compared to regular decoding, \ours achieves 44$\times$ FLOP/s and 41$\times$ Operational Intensity under the setting of batch size 16 and sequence length 1024 with 64 candidate tokens.
 Fig.~\ref{fig:llama33b-spec-seq1024} and Table~\ref{tab:llama33b-spec-seq1024} illustrate the effects of \ours decoding while keeping the sequence length fixed at 1024. Increasing the batch size does not improve Operational Intensity in this scenario. 

\begin{figure}[h]
    \centering
    \includegraphics[width=0.8\textwidth]{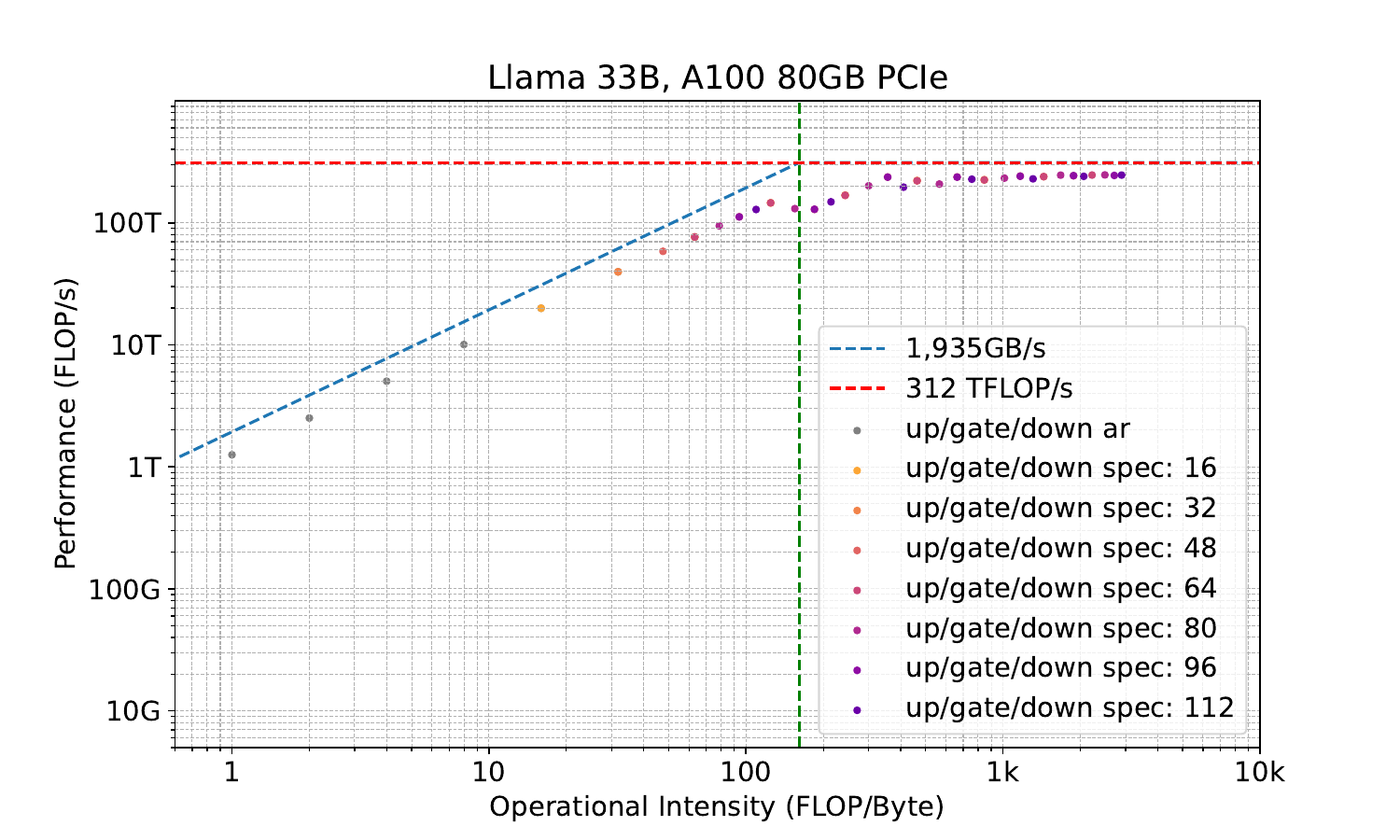}
    \caption{FLOP/s vs. Operational Intensity of Linear layers.}
    \label{fig:llama33b-spec--mlp-bsall}
\end{figure}

Next, we examine the linear layer, focusing on the up/gate/down linear layers. The results are shown in Fig.~\ref{fig:llama33b-spec--mlp-bsall} and Table~\ref{tab:llama33b-spec--mlp-bsall}. \textcolor{black}{Since the linear layers in the decoding phase only process the future tokens while the past tokens are cached, they are independent of the sequence length.} We vary the batch size to observe the effects. As \ours increases the number of candidate tokens with the increasing batch size, we observe a shift from a memory-bandwidth-bound region to a computation-bound region. This shift demonstrates how \ours can transition the performance characteristics of the linear layers from being limited by memory bandwidth to being limited by computational capacity.

\clearpage

\begin{table}[h]
\centering
\scriptsize
\begin{tabular}{lcccccccc}

\toprule
Seq. Length & \multicolumn{8}{c}{Number of Candidate Tokens} \\
\midrule
 & 1 & 16 & 32 & 48 & 64 & 80 & 96 & 112 \\
\midrule
128  & 0.54 \& 0.98 & 7.87 \& 12.8 & 14.73 \& 21.33 & 19.78 \& 27.43 & 25.25 \& 32.0 & 28.63 \& 35.56 & 32.58 \& 38.4 & 36.57 \& 40.73 \\
256  & 0.75 \& 0.99 & 11.2 \& 13.47 & 21.29 \& 23.27 & 28.69 \& 30.72 & 36.59 \& 36.57 & 41.2 \& 41.29 & 45.99 \& 45.18 & 52.33 \& 48.43 \\
512  & 1.02 \& 0.99 & 14.69 \& 13.84 & 27.47 \& 24.38 & 37.35 \& 32.68 & 47.09 \& 39.38 & 52.24 \& 44.91 & 59.55 \& 49.55 & 66.35 \& 53.49 \\
1024  & 1.24 \& 0.99 & 17.42 \& 14.03 & 32.15 \& 24.98 & 43.89 \& 33.76 & 54.8 \& 40.96 & 60.19 \& 46.97 & 68.28 \& 52.07 & 75.45 \& 56.44 \\
2048 & 1.39 \& 0.99 & 19.03 \& 14.12 & 35.05 \& 25.28 & 48.03 \& 34.32 & 59.66 \& 41.8 & 63.91 \& 48.08 & 72.83 \& 53.43 & 80.05 \& 58.04 \\
4096 & 1.48 \& 0.99 & 19.8 \& 14.17 & 36.59 \& 25.44 & 50.4 \& 34.61 & 62.29 \& 42.23 & 65.84 \& 48.65 & 74.86 \& 54.13 & 82.06 \& 58.87 \\
8192 & 1.53 \& 0.99 & 20.08 \& 14.2 & 36.89 \& 25.52 & 50.44 \& 34.76 & 62.11 \& 42.45 & 67.5 \& 48.94 & 76.97 \& 54.49 & 84.5 \& 59.3 \\
\bottomrule
\end{tabular}
\caption{
TFLOP/s \& Operational Intensity of attention matrix multiplication with batch size 16 for Llama 33B on an A100 80GB PCIe.}
\label{tab:llama33b-spec-bs16}
\end{table}

\begin{table}[h]
\centering
\scriptsize
\begin{tabular}{lcccccccc}
\toprule
Batch Size & \multicolumn{8}{c}{Number of Candidate Tokens} \\
\midrule
 & 1 & 16 & 32 & 48 & 64 & 80 & 96 & 112 \\
\midrule
1  & 0.37 \& 0.99 & 5.22 \& 14.03 & 10.15 \& 24.98 & 15.02 \& 33.76 & 19.79 \& 40.96 & 21.52 \& 46.97 & 25.65 \& 52.07 & 29.4 \& 56.44 \\
2  & 0.54 \& 0.99 & 8.25 \& 14.03 & 16.0 \& 24.98 & 21.62 \& 33.76 & 28.24 \& 40.96 & 31.84 \& 46.97 & 37.49 \& 52.07 & 43.04 \& 56.44 \\
4  & 0.75 \& 0.99 & 11.41 \& 14.03 & 21.97 \& 24.98 & 30.02 \& 33.76 & 38.71 \& 40.96 & 43.41 \& 46.97 & 50.06 \& 52.07 & 56.77 \& 56.44 \\
8  & 1.02 \& 0.99 & 14.78 \& 14.03 & 27.78 \& 24.98 & 38.09 \& 33.76 & 47.99 \& 40.96 & 53.32 \& 46.97 & 61.0 \& 52.07 & 68.11 \& 56.44 \\
16 & 1.24 \& 0.99 & 17.42 \& 14.03 & 32.15 \& 24.98 & 43.89 \& 33.76 & 54.8 \& 40.96 & 60.19 \& 46.97 & 68.28 \& 52.07 & 75.45 \& 56.44 \\
32 & 1.39 \& 0.99 & 18.89 \& 14.03 & 34.67 \& 24.98 & 47.57 \& 33.76 & 58.89 \& 40.96 & 63.61 \& 46.97 & 72.17 \& 52.07 & 79.21 \& 56.44 \\
64 & 1.48 \& 0.99 & 19.58 \& 14.03 & 35.87 \& 24.98 & 49.45 \& 33.76 & 61.13 \& 40.96 & 64.84 \& 46.97 & 73.73 \& 52.07 & 81.02 \& 56.44 \\

\bottomrule
\end{tabular}
\caption{
TFLOP/s \& Operational Intensity of attention matrix multiplication with sequence length 1024 for Llama 33B on an A100 80GB PCIe.}
\label{tab:llama33b-spec-seq1024}
\end{table}

\begin{table}[h]
\centering
\tiny
\begin{tabular}{lcccccccc}
\toprule
Batch Size & \multicolumn{8}{c}{Number of Candidate Tokens} \\
\midrule
 & 1 & 16 & 32 & 48 & 64 & 80 & 96 & 112 \\
\midrule
1  & 1.26 \& 1.0 & 19.95 \& 15.95 & 39.69 \& 31.79 & 58.4 \& 47.53 & 76.57 \& 63.17 & 94.4 \& 78.7 & 111.91 \& 94.14 & 128.64 \& 109.47 \\
2  & 2.51 \& 2.0 & 39.66 \& 31.79 & 76.53 \& 63.17 & 112.05 \& 94.14 & 145.73 \& 124.71 & 130.67 \& 154.89 & 129.1 \& 184.69 & 148.56 \& 214.12 \\
4  & 5.03 \& 4.0 & 76.44 \& 63.17 & 145.8 \& 124.71 & 128.85 \& 184.69 & 167.85 \& 243.17 & 201.19 \& 300.21 & 236.93 \& 355.85 & 195.91 \& 410.14 \\
8  & 10.06 \& 7.99 & 145.72 \& 124.71 & 168.26 \& 243.17 & 236.83 \& 355.85 & 221.11 \& 463.14 & 207.79 \& 565.44 & 236.95 \& 663.07 & 227.8 \& 756.36 \\
16 & 19.96 \& 15.95 & 168.35 \& 243.17 & 221.41 \& 463.14 & 237.5 \& 663.07 & 224.71 \& 845.59 & 232.49 \& 1012.87 & 241.12 \& 1166.74 & 229.25 \& 1308.76 \\
32 & 39.69 \& 31.79 & 221.74 \& 463.14 & 224.88 \& 845.59 & 241.33 \& 1166.74 & 239.02 \& 1440.25 & 245.83 \& 1675.97 & 243.55 \& 1881.24 & 240.33 \& 2061.59 \\
64 & 76.57 \& 63.17 & 225.19 \& 845.59 & 239.2 \& 1440.25 & 243.26 \& 1881.24 & 246.16 \& 2221.31 & 246.91 \& 2491.55 & 244.52 \& 2711.46 & 246.14 \& 2893.91 \\
\bottomrule
\end{tabular}
\caption{
TFLOP/s \& Operational Intensity of linear layers (up/gate/down) for Llama 33B on an A100 80GB PCIe.
}\label{tab:llama33b-spec--mlp-bsall}
\end{table}

\subsection{Predicting \ours Performance}

\begin{figure}[h]
    \centering
    \includegraphics[width=0.8\textwidth]{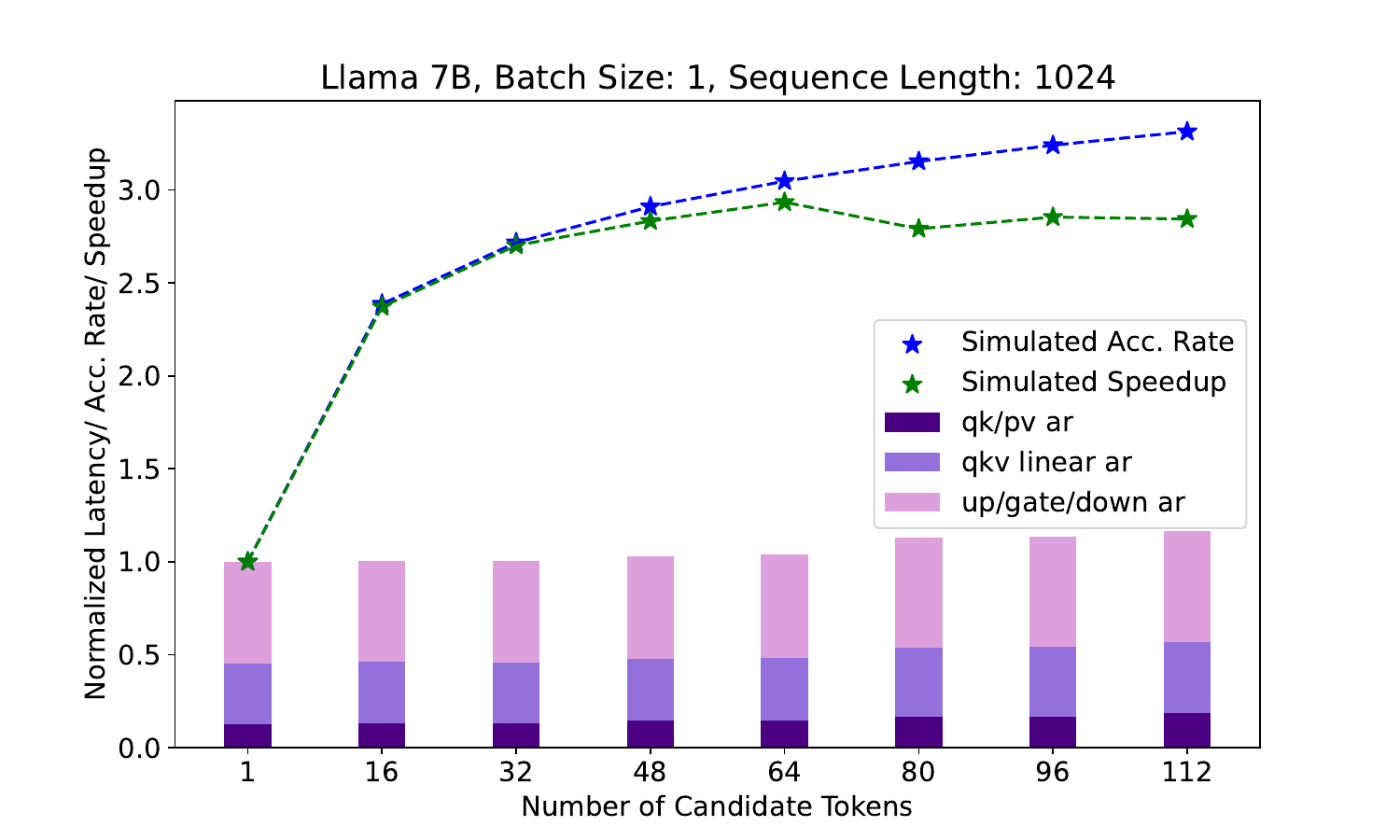}
    \caption{Simulated acceleration rate, speedup, and normalized latency ablation using different numbers of candidate tokens under the setting of batch size 1 and sequence length 1024 for Llama-7B on an A100 80GB PCIe.}
    \label{fig:llama7b-sim-bs1-seq1024}
\end{figure}

We further employ a straightforward analytical model \textcolor{black}{for} the acceleration rate. The ablation study results in Sec.~\ref{section:config of tree} indicate that the acceleration rate can be approximated by a simple logarithmic function. Using the results from Fig.~\ref{fig:sparse_acc}, we model the curve as $\texttt{acc\_rate} = 0.477 \log(\texttt{num\_candidate})$. We simulate the latency of one simplified block of the Llama-7B model (sequentially processing $XW_Q$, $XW_K$, $XW_V$, $QK^T$, $PV$, $XW_u$, $XW_g$, $XW_d$) by first fixing the batch size at 1 and the sequence length at 1024.
\textcolor{black}{
The candidate tokens are processed parallelly by constructing the tree attention described in Section~\ref{sec:tree_attention}. We omit the latency of the post-processing steps including verification and acceptance for \ours since they introduce marginal overhead.
}
Fig.~\ref{fig:llama7b-sim-bs1-seq1024} illustrates the simulated acceleration rate and speedup for different numbers of candidate tokens under these settings. As the number of candidate tokens increases, both the acceleration rate and speedup initially show improvements. However, beyond 64, the speedup starts to decline, indicating diminishing returns with further increases in candidate length. This aligns with the experimental results in Fig.~\ref{fig:sparse_speed} and suggests that there is an optimal range for the numbers of candidate tokens where \ours provides the most significant performance gains.

\begin{figure}[h]
    \centering
    \includegraphics[width=0.8\textwidth]{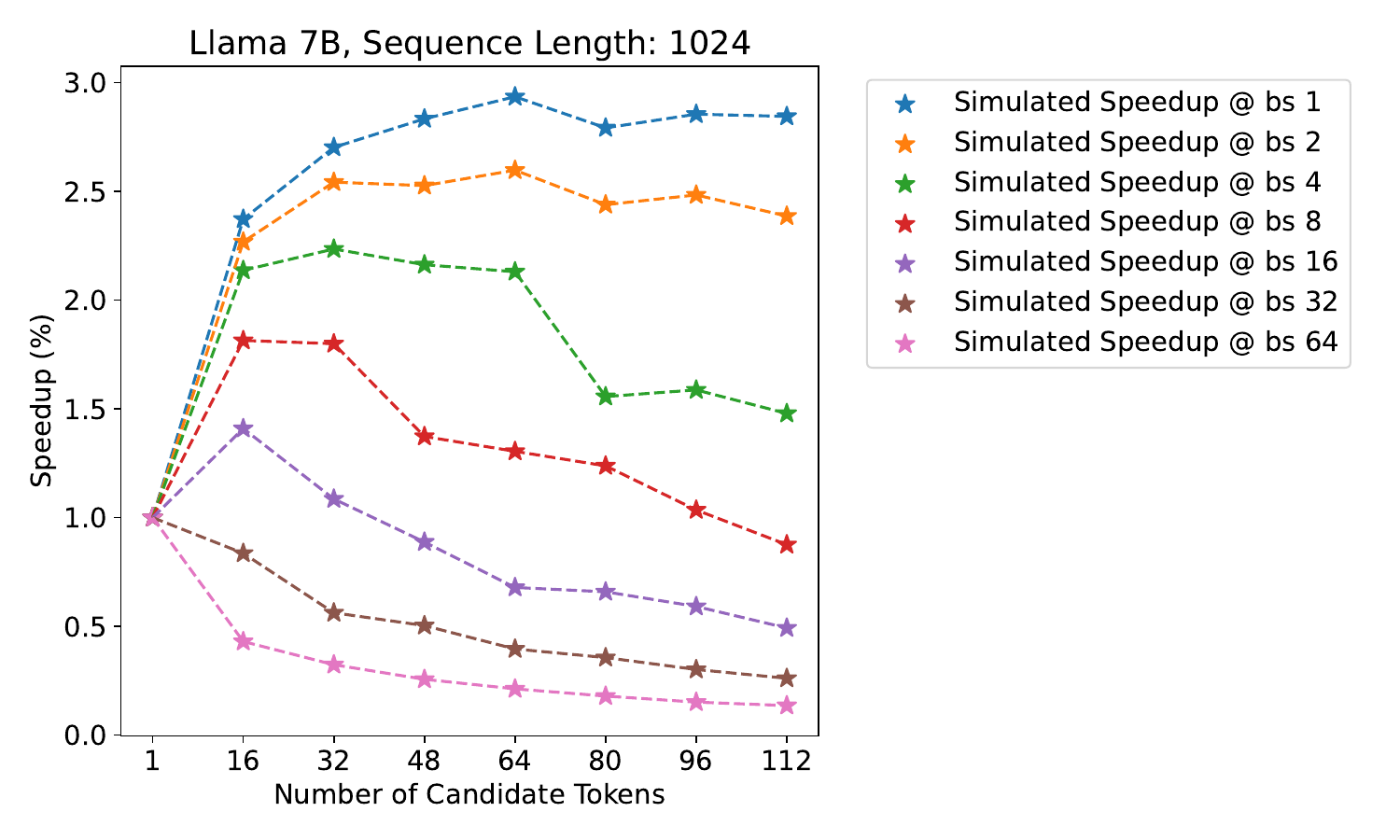}
    \caption{Simulated speedup with sequence length 1024 for Llama-7B.}
    \label{fig:llama7b-sim-bs1-allbs}
\end{figure}

We plot the simulated speedup under different batch size settings with a fixed sequence length of 1024 in Fig.~\ref{fig:llama7b-sim-bs1-allbs}. The results indicate that when the batch size exceeds 32, the speedup decreases and may even have a negative effect. This occurs because the linear layers shift from being memory-bandwidth-bound to computationally bound.

\begin{figure}[h]
    \centering
    \includegraphics[width=0.8\textwidth]{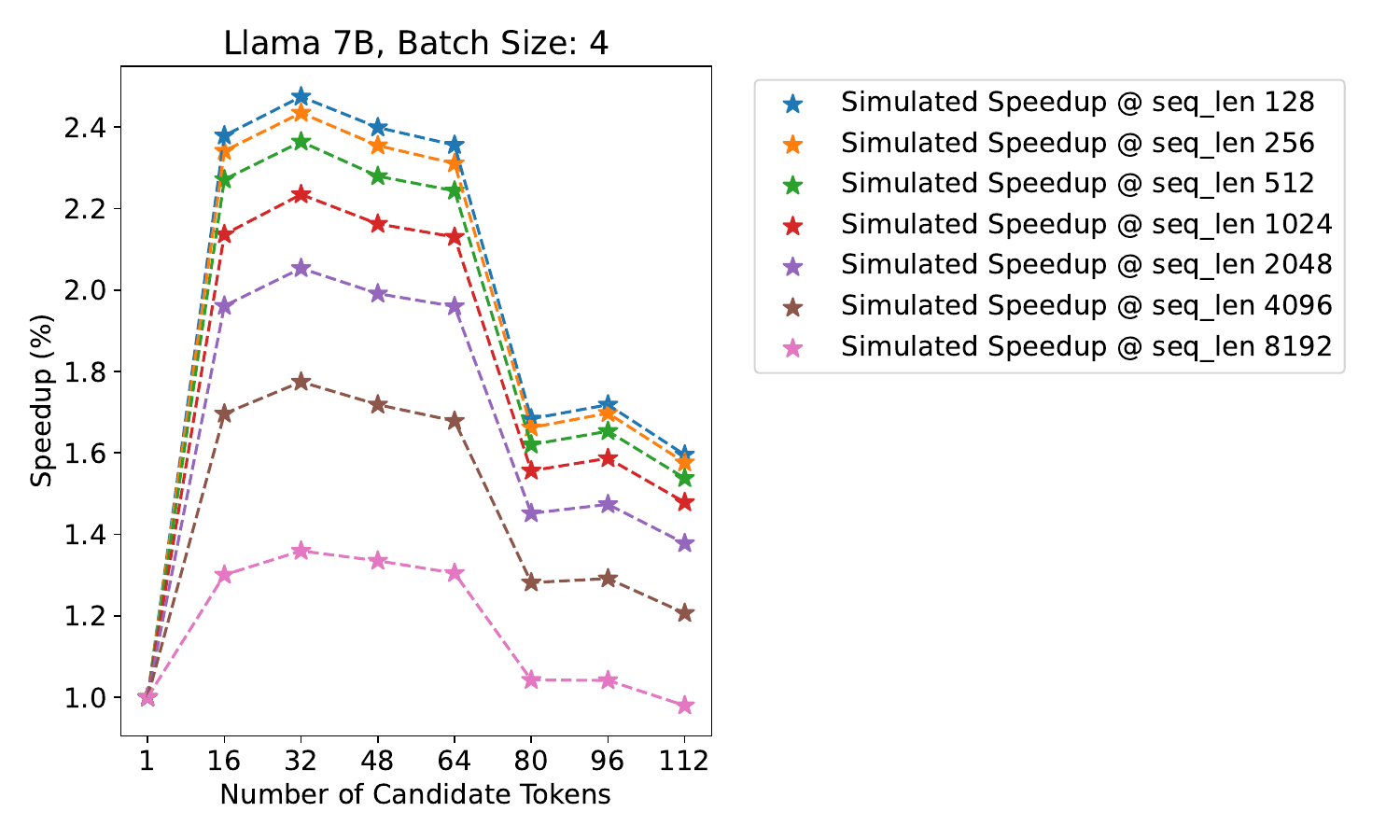}
    \caption{Simulated speedup with batch size 4 for Llama-7B.}
    \label{fig:llama7b-sim-allseq}
\end{figure}

We conduct another experiment using a batch size of 4 and different sequence lengths. As shown in Fig.~\ref{fig:llama7b-sim-allseq}, the optimal number of candidate tokens remains relatively consistent across different sequence lengths. However, as the sequence length increases, the overall performance decreases. This performance drop is primarily due to the overhead from attention matrix multiplication, while the linear layer computation remains constant \textcolor{black}{since the computation of linear layers is independent of the sequence length.}

Our simulations show that the optimal number of candidate tokens is key for model scaling with \ours, as benefits decrease beyond a certain range. Initially, increasing batch size improves performance through parallelism, but too large a batch size shifts linear layers from memory-bandwidth-bound to compute-bound, reducing speedup. Longer sequences increase attention matrix multiplication overhead, lowering performance, and emphasizing the need to optimize attention mechanisms. Effective model scaling requires balancing the number of candidate tokens, adjusting batch sizes to avoid compute-bound transitions, and enhancing attention mechanisms for longer sequences. These strategies ensure better resource utilization and higher performance, demonstrating the value of simulations in predicting performance and guiding acceleration strategy design.

\end{document}